\documentclass[10pt,journal,compsoc]{IEEEtran}
\ifCLASSOPTIONcompsoc
  \usepackage[nocompress]{cite}
\else
  \usepackage{cite}
\fi
\ifCLASSINFOpdf
\else
\fi
\usepackage{graphicx}
\usepackage{amsmath}
\usepackage{amssymb}
\usepackage{booktabs}
\usepackage{makecell}
\usepackage{diagbox}
\usepackage{algorithm}
\usepackage{algorithmic}
\usepackage[flushleft]{threeparttable}
\usepackage{multirow}
\usepackage{url}
\usepackage{color}

\usepackage{amsmath,amsfonts,bm}

\def\Figref#1{Figure~\ref{#1}}

\def\eqref#1{equation~\ref{#1}}
\def\1{\bm{1}}

\def\vf{{\bm{f}}}

\def\vx{{\bm{x}}}

\def\mM{{\bm{M}}}

\def\mU{{\bm{U}}}

\DeclareMathAlphabet{\mathsfit}{\encodingdefault}{\sfdefault}{m}{sl}
\SetMathAlphabet{\mathsfit}{bold}{\encodingdefault}{\sfdefault}{bx}{n}

\def\sC{{\mathbb{C}}}
\def\sD{{\mathbb{D}}}
\def\sQ{{\mathbb{Q}}}

\def\sS{{\mathbb{S}}}

\def\sX{{\mathbb{X}}}
\def\sY{{\mathbb{Y}}}
\def\sZ{{\mathbb{Z}}}

\begin{document}
%
% paper title
% Titles are generally capitalized except for words such as a, an, and, as,
% at, but, by, for, in, nor, of, on, or, the, to and up, which are usually
% not capitalized unless they are the first or last word of the title.
% Linebreaks \\ can be used within to get better formatting as desired.
% Do not put math or special symbols in the title.
% \title{MahiNet: A Neural Network for Many-Class Few-Shot Learning with Class Hierarchy}
\title{Many-Class Few-Shot Learning on Multi-Granularity Class Hierarchy}
%
%
% author names and IEEE memberships
% note positions of commas and nonbreaking spaces ( ~ ) LaTeX will not break
% a structure at a ~ so this keeps an author's name from being broken across
% two lines.
% use \thanks{} to gain access to the first footnote area
% a separate \thanks must be used for each paragraph as LaTeX2e's \thanks
% was not built to handle multiple paragraphs
%
%
%\IEEEcompsocitemizethanks is a special \thanks that produces the bulleted
% lists the Computer Society journals use for "first footnote" author
% affiliations. Use \IEEEcompsocthanksitem which works much like \item
% for each affiliation group. When not in compsoc mode,
% \IEEEcompsocitemizethanks becomes like \thanks and
% \IEEEcompsocthanksitem becomes a line break with idention. This
% facilitates dual compilation, although admittedly the differences in the
% desired content of \author between the different types of papers makes a
% one-size-fits-all approach a daunting prospect. For instance, compsoc 
% journal papers have the author affiliations above the "Manuscript
% received ..."  text while in non-compsoc journals this is reversed. Sigh.

\author{Lu~Liu, Tianyi Zhou,
        Guodong Long, Jing Jiang
        and~Chengqi Zhang% <-this % stops a space
\IEEEcompsocitemizethanks{
\IEEEcompsocthanksitem Corresponding author: Guodong Long
\IEEEcompsocthanksitem Lu Liu, Guodong Long, Jing Jiang and Chengqi Zhang are with Centre for Artificial Intelligence, FEIT, University of Technology Sydney, Ultimo, NSW 2007, Australia.\protect\\
% note need leading \protect in front of \\ to get a newline within \thanks as
% \\ is fragile and will error, could use \hfil\break instead.
E-mails: lu.liu-10@student.uts.edu.au, guodong.long@uts.edu.au, jing.jiang@uts.edu.au, chengqi.zhang@uts.edu.au

\IEEEcompsocthanksitem
Tianyi Zhou is with Paul G. Allen School of Computer Science \& Engineering, University of Washington, Seattle, WA 98195, USA.\protect\\ 
Email: tianyizh@uw.edu}% <-this % stops an unwanted space
% \thanks{Manuscript received April 19, 2005; revised August 26, 2015.}
}

% note the % following the last \IEEEmembership and also \thanks - 
% these prevent an unwanted space from occurring between the last author name
% and the end of the author line. i.e., if you had this:
% 
% \author{....lastname \thanks{...} \thanks{...} }
%                     ^------------^------------^----Do not want these spaces!
%
% a space would be appended to the last name and could cause every name on that
% line to be shifted left slightly. This is one of those "LaTeX things". For
% instance, "\textbf{A} \textbf{B}" will typeset as "A B" not "AB". To get
% "AB" then you have to do: "\textbf{A}\textbf{B}"
% \thanks is no different in this regard, so shield the last } of each \thanks
% that ends a line with a % and do not let a space in before the next \thanks.
% Spaces after \IEEEmembership other than the last one are OK (and needed) as
% you are supposed to have spaces between the names. For what it is worth,
% this is a minor point as most people would not even notice if the said evil
% space somehow managed to creep in.

% The paper headers
\markboth{Accepted to IEEE Transactions on Knowledge and Data Engineering}%
{Shell \MakeLowercase{\textit{et al.}}: Bare Demo of IEEEtran.cls for Computer Society Journals}
% The only time the second header will appear is for the odd numbered pages
% after the title page when using the twoside option.
% 
% *** Note that you probably will NOT want to include the author's ***
% *** name in the headers of peer review papers.                   ***
% You can use \ifCLASSOPTIONpeerreview for conditional compilation here if
% you desire.

% The publisher's ID mark at the bottom of the page is less important with
% Computer Society journal papers as those publications place the marks
% outside of the main text columns and, therefore, unlike regular IEEE
% journals, the available text space is not reduced by their presence.
% If you want to put a publisher's ID mark on the page you can do it like
% this:
%\IEEEpubid{0000--0000/00\$00.00~\copyright~2015 IEEE}
% or like this to get the Computer Society new two part style.
%\IEEEpubid{\makebox[\columnwidth]{\hfill 0000--0000/00/\$00.00~\copyright~2015 IEEE}%
%\hspace{\columnsep}\makebox[\columnwidth]{Published by the IEEE Computer Society\hfill}}
% Remember, if you use this you must call \IEEEpubidadjcol in the second
% column for its text to clear the IEEEpubid mark (Computer Society jorunal
% papers don't need this extra clearance.)

% use for special paper notices
%\IEEEspecialpapernotice{(Invited Paper)}

% for Computer Society papers, we must declare the abstract and index terms
% PRIOR to the title within the \IEEEtitleabstractindextext IEEEtran
% command as these need to go into the title area created by \maketitle.
% As a general rule, do not put math, special symbols or citations
% in the abstract or keywords.
\IEEEtitleabstractindextext{%
\begin{abstract}
We study many-class few-shot (MCFS) problem in both supervised learning and meta-learning settings. 
Compared to the well-studied many-class many-shot and few-class few-shot problems, the MCFS problem commonly occurs in practical applications but has been rarely studied in previous literature. It brings new challenges of distinguishing between many classes given only a few training samples per class.
In this paper, we leverage the class hierarchy as a prior knowledge to train a coarse-to-fine classifier that can produce accurate predictions for MCFS problem in both settings. 
%we address MCFS problem by solving a multi-output learning problem.
%Specifically, 
The propose model, ``memory-augmented hierarchical-classification network (MahiNet)'', performs coarse-to-fine classification where each coarse class can cover multiple fine classes.
Since it is challenging to directly distinguish a variety of fine classes given few-shot data per class, MahiNet starts from learning a classifier over coarse-classes with more training data whose labels are much cheaper to obtain. 
The coarse classifier reduces the searching range over the fine classes and thus alleviates the challenges from ``many classes''.
On architecture, MahiNet firstly deploys a convolutional neural network (CNN) to extract features. It then integrates a memory-augmented attention module and a multi-layer perceptron (MLP) together to produce the probabilities over coarse and fine classes.
While the MLP extends the linear classifier, the attention module extends the KNN classifier, both together targeting the ``few-shot'' problem.
We design several training strategies of MahiNet for supervised learning and meta-learning.
In addition, we propose two novel benchmark datasets ``\textit{mcfs}ImageNet'' (as a subset of ImageNet) and ``\textit{mcfs}Omniglot'' (re-splitted Omniglot) specially designed for MCFS problem.
In experiments, we show that MahiNet outperforms several state-of-the-art models (e.g., prototypical networks and relation networks) on MCFS problems in both supervised learning and meta-learning.
\end{abstract}

% Note that keywords are not normally used for peerreview papers.
}

% make the title area
\maketitle

% To allow for easy dual compilation without having to reenter the
% abstract/keywords data, the \IEEEtitleabstractindextext text will
% not be used in maketitle, but will appear (i.e., to be "transported")
% here as \IEEEdisplaynontitleabstractindextext when the compsoc 
% or transmag modes are not selected <OR> if conference mode is selected 
% - because all conference papers position the abstract like regular
% papers do.
\IEEEdisplaynontitleabstractindextext
% \IEEEdisplaynontitleabstractindextext has no effect when using
% compsoc or transmag under a non-conference mode.

% For peer review papers, you can put extra information on the cover
% page as needed:
% \ifCLASSOPTIONpeerreview
% \begin{center} \bfseries EDICS Category: 3-BBND \end{center}
% \fi
%
% For peerreview papers, this IEEEtran command inserts a page break and
% creates the second title. It will be ignored for other modes.
\IEEEpeerreviewmaketitle

\IEEEraisesectionheading{\section{Introduction}\label{sec:introduction}}
% Computer Society journal (but not conference!) papers do something unusual
% with the very first section heading (almost always called "Introduction").
% They place it ABOVE the main text! IEEEtran.cls does not automatically do
% this for you, but you can achieve this effect with the provided
% \IEEEraisesectionheading{} command. Note the need to keep any \label that
% is to refer to the section immediately after \section in the above as
% \IEEEraisesectionheading puts \section within a raised box.

% The very first letter is a 2 line initial drop letter followed
% by the rest of the first word in caps (small caps for compsoc).
% 
% form to use if the first word consists of a single letter:
% \IEEEPARstart{A}{demo} file is ....
% 
% form to use if you need the single drop letter followed by
% normal text (unknown if ever used by the IEEE):
% \IEEEPARstart{A}{}demo file is ....
% 
% Some journals put the first two words in caps:
% \IEEEPARstart{T}{his demo} file is ....
% 
% Here we have the typical use of a "T" for an initial drop letter
% and "HIS" in caps to complete the first word.
\IEEEPARstart{T}{he} representation power of deep neural networks (DNN) has significantly improved in recent years, as deeper, wider and more complicated DNN architectures have emerged to match the increasing computation power of new hardware~\cite{He_2016_CVPR,huang2017densely}.
Although this brings hope to complex tasks that could be hardly solved by previous shallow models, more labeled data is usually required to train the deep models.
The scarcity of annotated data has become a new bottleneck for training more powerful DNNs. It is quite common in practical applications such as image search, robot navigation and video surveillance. For example, in image classification, the number of candidate classes easily exceeds tens of thousands (i.e., many-class), but the training samples available for each class can be less than $100$ (i.e., few-shot). Unfortunately, this scenario is beyond of the scope of current meta-learning methods for few-shot classification, which aims to address the data scarcity (few-shot data per class) but the number of classes in each task is usually less than 10. Additionally, in life-long learning, models are always updated once new training data becomes available, and those models are expected to quickly adapt to new classes with a few training samples available.

\begin{table}
\setlength{\tabcolsep}{5.5pt}
\centering
\caption{Targeted problems of different methods. MahiNet targets many-class few-shot problem, which is more challenging and practical than others.}
% \vspace{-0.5em}
\begin{tabular}{|l|c|c|}
\hline
& many-class & few-class \\ \hline
few-shot   & \textbf{MahiNet (ours)}    & MAML, Matching Net, \emph{etc.} \\ \hline
many-shot  & \multicolumn{2}{c|}{ResNet, DenseNet, Inception, VGG, \emph{etc.}}   \\ 
\hline
\end{tabular}
\label{table:alo-setting}
% \vspace{-1.5em}
\end{table}

\begin{figure*}[t!]
\begin{center}
\includegraphics[width=\linewidth]{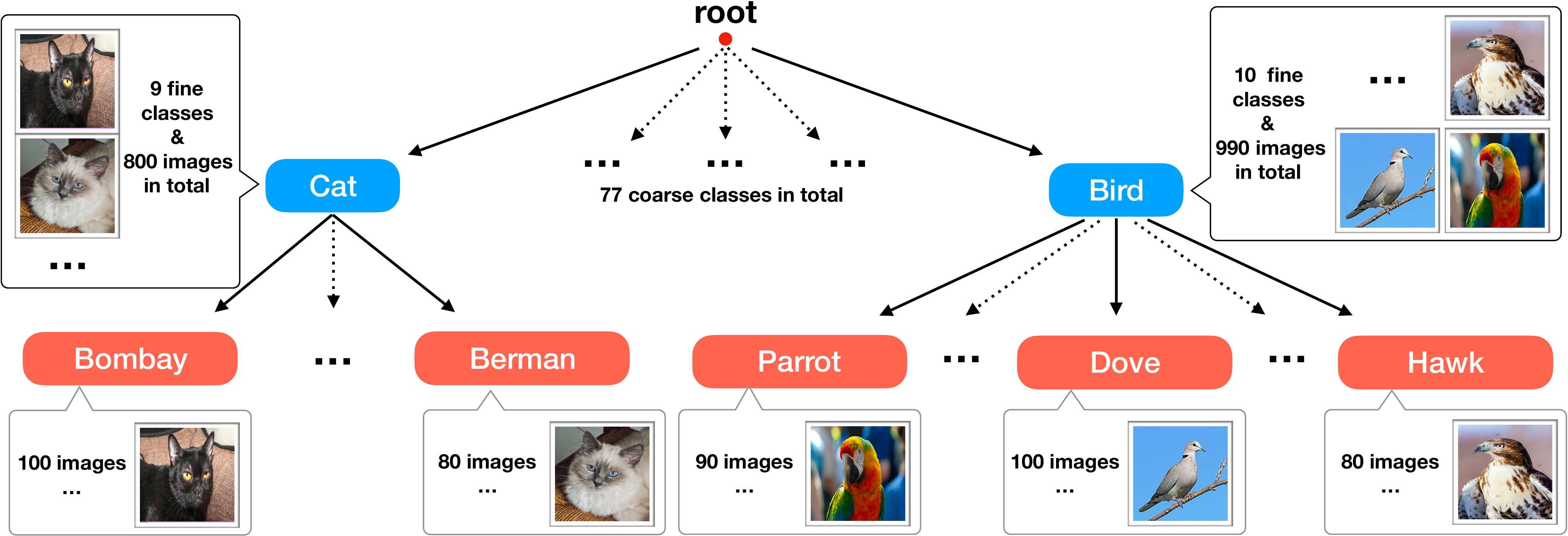}
\end{center}
\caption{
%Many-class few-shot learning (MCFS) problem with class hierarchy information.
A many-class few-shot learning (MCFS) problem using multi-label class hierarchy information.
There are a few coarse classes (blue) and each coarse class covers a large number of fine classes (red) so that the total number of fine classes is large. Only a few training samples are available for each fine class. The goal is to train a classifier to generate a prediction over all fine classes with the help of coarse prediction.
We utilize meta-learning to solve the problem of many-class few-shot learning problem, where each task is an MCFS problem sampled from a certain distribution. The meta-learner's goal is to help train a classifier for any sampled task with better adaptation to few-shot data within the sampled task.
}
\label{fig:data-example}
\end{figure*}

Although previous works of fully supervised learning have shown the remarkable power of DNN when ``many-class many-shot'' training data is available, their performance degrades dramatically when each class only has a few samples available for training.
In practical applications, acquiring samples of rare species or personal data from edge devices is usually difficult, expensive, and forbidden due to privacy protection. 
% Uploading personal data to the cloud is a way to accumulate big data but has issues with users' privacy.
In many-class cases, annotating even one additional sample per class can be very expensive and requires a lot of human efforts. Moreover, the training set cannot be fully balanced over all the classes in practice.
In these few-shot learning scenarios, the capacity of a deep model cannot be fully utilized, and it becomes much harder to generalize the model to unseen data. Recently, several approaches have been proposed to address the few-shot learning problem. Most of them are based on the idea of ``meta-learning'', which trains a meta-learner over different few-shot tasks so it can generalize to new few-shot tasks with unseen classes.
Thereby, the meta-learner aims to learn a stronger prior encoding the general knowledge achieved during learning various tasks, so it is capable to help a learner model quickly adapt to a new task with new classes of insufficient training samples.
Meta-learning can be categorized into two types: methods based on ``learning to optimize'', and methods based on metric learning. The former type adaptively modifies the optimizer (or some parts of it) used for training the task-specific model. It includes methods that incorporate an Recurrent Neural Networks (RNN) meta-learner~\cite{metaRNN, learn2opt, ravi2017optimization}, and model-agnostic meta-learning (MAML) methods aiming to learn a generally compelling initialization~\cite{finn2017model}. The second type learns a similarity/distance metric~\cite{vinyals2016matching} or a model generating a support set of samples~\cite{snell2017prototypical} that can be used to build K-Nearest Neighbors (KNN) classifiers from few-shot data in different tasks. 

Instead of meta-learning methods, data augmentation based approaches, such as the hallucination method proposed in \cite{douze2018low}, address the few-shot learning problem by generating more artificial samples for each class. 
However, most existing few-shot learning approaches only focus on ``few-class'' cases (e.g., $5$ or $10$) per task, and their performance drastically collapses when the number of classes slightly grows to tens to hundreds. This is because the samples per class no longer provide enough information to distinguish them from other possible samples within a large number of other classes. And in real-world few-shot problems, a task is usually complicated involving many classes.

Fortunately, in practice, multi-outputs/labels information such as coarse-class labels in a class hierarchy is usually available or cheaper to obtain. In this case, the correlation between fine and coarse classes can be leveraged in solving the MCFS problem, e.g., by training a coarse-to-fine hierarchical prediction model.
As shown in Fig.~\ref{fig:data-example}, coarse class labels might reveal the relationships among the targeted fine classes. Moreover, the samples per coarse class are sufficient to train a reliable coarse classifier, whose predictions are able to narrow down the candidates for the corresponding fine class.
For example, a sheepdog with long hair could be easily mis-classified as a mop when training samples of sheepdog are insufficient.
However, if we could train a reliable dog classifier, it would be much simpler to predict an image as a sheepdog than a mop given a correct prediction of the coarse class as ``dog''.
Training coarse-class prediction models is much easier and less suffered from the ``many class few shot''problem (since the coarse classes are fewer than the fine classes and the samples per coarse class are much more than that for each fine class). It provides helpful information to fine-class prediction due to the relationship between coarse classes and fine classes.
Hence, class hierarchy might provide weakly supervised information to help solve the ``many-class few-shot (MCFS)'' problem in a framework of multi-output learning, which aims to predict the class label on each level of the class hierarchy.

% \subsection{Our Approach}

In this paper, we address the MCFS problem in both traditional supervised learning and in meta-learning settings by exploring the multi-output information on multiple class hierarchy levels.
We develop a neural network architecture ``memory-augmented hierarchical-classification networks (MahiNet)'' that can be applied to both learning settings. 
MahiNet uses a Convolutional Neural Network (CNN), i.e., ResNet~\cite{He_2016_CVPR}, as a backbone network to firstly extract features from raw images. It then trains coarse-class and fine-class classifiers based on the features, and combines their outputs to produce the final probability prediction over the fine classes.
In this way, both the coarse-class and the fine-class classifiers mutually help each other within MahiNet: the coarse classifier helps to narrow down the candidates for the fine classifier, while the fine classifier provides multiple attributes describing every coarse class and can regularize the coarse classifier. This design leverages the relationship between the fine classes as well as the relationship between the fine classes and the coarse classes, which mitigates the difficulty caused by the ``many class'' challenge. To the best of our knowledge, we are the first to successfully train a multi-output model to employ existing class hierarchy prior for improving both few-shot learning and supervised learning. It is different from those methods discussed in related works that use a latent clustering hierarchy instead of the explicit class hierarchy we use.
In Table~\ref{table:alo-setting}, we provide a brief comparison of MahiNet with other popular models on the learning scenarios they excel.

To address the ``few-shot'' problem, we apply two types of classifiers in MahiNet, i.e., MLP classifier and KNN classifier, which respectively have advantages in many-shot and few-shot situations. 
In principle, we always use MLP for coarse classification, and KNN for fine classification. 
Specially, with a sufficient amount of data in supervised learning, MLP is combined with KNN for fine classification; and in meta-learning when less data is available, we also use KNN for coarse classification to assist MLP.

To make the KNN learnable and adaptive to new classes with few-shot data, we train an attention module to provide the similarity/distance metric used in KNN, and a re-writable memory of limited size to store and update the KNN support set during training. In supervised learning, it is necessary to maintain and update a relatively small memory (e.g., 7.2\% of the dataset in our experiment) by selecting a few representative samples, because conducting a KNN search over all available training samples is too expensive in computation. In meta-learning, the attention module can be treated as a meta-learner that learns a universal similarity metric across different tasks.

Since the commonly used datasets in meta-learning do not have hierarchical multi-label annotations, 
we randomly extract a large subset of ImageNet~\cite{imagenet}  ``\textit{mcfs}ImageNet'' as a benchmark dataset specifically designed for MCFS problem. 
Each sample in \textit{mcfs}ImageNet has two labels: a coarse class label and a fine class label.
It contains 139,346 images from $77$ non-overlapping coarse classes composed of $754$ randomly sampled fine classes, each has only $\approx 180$ images, which might be further splitted for training, validation and test.
The imbalance between different classes in the original ImageNet are preserved to reflect the imbalance in practical problems. 
Similarly, we further extract ``\textit{mcfs}Omniglot'' from Omniglot~\cite{lake2011one} for the same purpose.
We will make them publicly available later.
In fully supervised learning experiments on these two datasets, MahiNet outperforms the widely used ResNet~\cite{He_2016_CVPR} (for fairness, MahiNet uses the same network (i.e., ResNet) as its backbone network).
In meta-learning scenario where each test task covers many classes, it shows more promising performance than popular few-shot methods including prototypical networks~\cite{snell2017prototypical} and relation networks~\cite{yang2018learning}, which are specifically designed for few-shot learning. 

Our contributions can be concluded as:
In the new version, we conclude our contributions at the end of introduction as follows:
Our contributions can be concluded as:
1) We address a new problem setting called ``many-class few-shot learning (MCFS)'' that has been widely encountered in practice. Compared to the conventional ``few-class few-shot (as in most few-shot learning methods)'' or ``many-class many-shot (as in most supervised learning methods)'' settings, MCFS is more practical and challenging but has been rarely studied in the ML community.
2) To alleviate the challenge of ``many-class few-shot'', we propose to utilize the knowledge of a predefined class hierarchy to train a model that takes the relationship between classes into account when generating a classifier over a relatively large number of classes.
3) To empirically justify whether our model can improve the MCFS problem by using class hierarchy information, we extract two new datasets from existing benchmarks, each coupled with a class hierarchy on reasonably specified classes. In experiments, we show that our method outperforms the other baselines in MCFS setting.

% 1) We shed the light on a problem called ``many-class few-shot''. Compared to conventional ``few-class few-shot'' or ``many-class many-shot'' problems, this targeted problem is more practical and challenging, but has been rarely studied.
% 2) To alleviate the challenge of ``many-class few-shot'', we propose to utilize the knowledge of predefined class hierarchy and a model designed to take the advantage of the provided hierarchy. 
% 3) To evaluate if our model can benefit from the predefined hierarchy, we propose two datasets with provided class hierarchy and show our model can outperform baselines on the problem of ``many-class few-shot''.

\section{Related Works}
\subsection{Few-shot Learning}
Generative models \cite{fei2006one} were trained to provide a global prior knowledge for solving the one-shot learning problem. With the advent of deep learning techniques, some recent approaches \cite{wong2015one, lake2013one} use generative models to encode specific prior knowledge, such as strokes and patches. More recently, works in \cite{douze2018low} and \cite{Wang-2018-105271} have applied hallucinations to training images and to generate more training samples, which converts a few-shot problem to a many-shot problem. 

Meta-learning has been widely studied to address the few-shot learning problems for fast adaptation to new tasks with only few-shot data. Meta-learning was first proposed in the last century \cite{naik1992meta, schmidhuber1987evolutionary}, and has recently brought significant improvements to few-shot learning, continual learning and online learning~\cite{zhao2019adaptive}. For example, the authors of~\cite{lake2015human} proposed a dataset of characters ``Omniglot'' for meta-learning while the work in \cite{koch2015siamese} apply a Siamese network to this dataset. A more challenging dataset ``\textit{mini}ImageNet''~\cite{ravi2017optimization,vinyals2016matching} was introduced later. \textit{mini}ImageNet is a subset of ImageNet~\cite{imagenet} and has more variety and higher recognition difficulty compared to Omniglot.
Researchers have also studied a combination of RNN and attention based method to overcome the few-shot problem~\cite{ravi2017optimization}.
More recently, the method in \cite{snell2017prototypical} was proposed based on metric learning to learn a shared KNN~\cite{Yu2019} classifier for different tasks. 
In contrast, the authors of \cite{finn2017model} developed their approach based on the second order optimization where the model can adapt quickly to new tasks or classes from an initialization shared by all tasks. The work in \cite{mishra2018simple} addresses the few-shot image recognition problem by temporal convolution, which sequentially encodes the samples in a task.
More recently, meta-learning strategy has been studied to solve other problems and tasks, such as using meta-learning training strategy to help design a more efficient unsupervised learning rule~\cite{metz2018learning}; mitigating the low-resource problems in neural machine translation task~\cite{gu2018meta}; alleviating the few annotations problem in object detection problem~\cite{schwartz2019repmet}. 

Our model is closely related to prototypical networks, in which every class is represented by a prototype that averages the embedding of all samples from that class, and the embedding is achieved by applying a shared encoder, i.e., the meta-learner.
It can be modified to generate an adaptive number of prototypes~\cite{allen2019infinite} and to handle extra weakly-supervised labels~\cite{ppn,liu2019learning} on a category graph.
Unlike these few-shot learning methods that produces a KNN classifier defined by the prototypes, our model produces multi-labels or multi-output predictions by combining the outputs of an MLP classifier and a KNN classifier, which are trained in either supervised learning or few-shot learning settings.

% In addition, our method targets many-class few-shot problem in both the supervised learning and the meta-learning settings.

\subsection{Multi-Label Classification}
Multi-label classification aims to assign multiple class labels to every sample~\cite{tsoumakas2007multi}. One solution is to use a chain of classifiers to turn multi-label problem into several binary classification problems~\cite{read2009classifier,read2014efficient,read2015scalable}.
Another solution treats multi-label classification as multi-class classification over all possible subsets of labels~\cite{spolaor2013comparison}. 
% Previous works solve multi-output problem by integrating different features under the theme of manifold regularization. 
%Other works learn an embedding or encoding space using different methods, e.g., cross-view learning~\cite{shen2018compact}, more efficient metric learning approaches~\cite{liu2018metric,zhang2005k}, feature-aware label space encoding, and label propagation.
Other works learn an embedding or encoding space using metric learning approaches~\cite{liu2018metric}, feature-aware label space encoding, label propagation, etc.
The work in~\cite{liu2017making} proposes a decision trees method for multi-label annotation problems.
The efficiency can be improved by using an ensemble of a pruned set of labels~\cite{read2008multi}.
The applications include multi-label classification for textual data~\cite{tsoumakas2009mining}, multi-label learning from crowds~\cite{8413163} and disease resistance prediction~\cite{heider2013multilabel}.
Our approach differs from these works in that the multi-output labels and predictions serve as auxiliary information to improve the many-class few-shot classification over fine classes.

\subsection{Hierarchical Classification}
Hierarchical classification is a special case of multi-label or multi-output problems~\cite{silla2011survey,gordon1987review}.
It has been applied to traditional supervised learning tasks~\cite{wehrmann2018hierarchical}, such as text classification~\cite{dumais2000hierarchical,sun2001hierarchical}, community data~\cite{gauch1981hierarchical}, case-based reasoning~\cite{8333767}, popularity prediction~\cite{yang2019hierarchical}, supergraph search in graph databases~\cite{lyu2019supergraph}, road networks~\cite{Ouyang18}, image annotation and robot navigation. However, to the best of our knowledge, our paper is the first work successfully leveraging class hierarchy information in few-shot learning and meta-learning tasks. Previous methods such as~\cite{ren2018meta}, have considered the class hierarchy information but failed to achieve improvement and thus did not integrate it in their method, whereas our method successfully leverage the hierarchical relationship between classes to improve the classification performance by using a memory-augmented model.

Similar idea of using hierarchy for few-shot learning has been studied  in~\cite{li2019large,yao2019hierarchically}, which learns to cluster semantic information and task representation, respectively.
\cite{ppn} utilizes coarsely labeled data in the hierarchy as weakly supervised data rather than multi-label annotations for few-shot learning.
In computational biology, hierarchy information has also been found helpful in gene function prediction, where two main taxonomies are Gene Ontology and Functional Catalogue. \cite{valentini2011true} proposed a truth path rule as an ensemble method to govern both taxonomies.
\cite{cesa2012synergy} shows that the key factors for the success of hierarchical ensemble methods are: (1)the integration and synergy among multi-label hierarchy, (2) data fusion, (3) cost-sensitive approaches, and (4) the strategy of selecting negative examples.
\cite{yu2015predicting} and~\cite{yu2018newgoa} address the incomplete annotations of proteins using label hierarchy (as studied in this paper) by following the idea of few-shot learning. Specifically, they learn to predict the new gene ontology annotations by Bi-random walks on a hybrid graph~\cite{yu2015predicting} and downward random walks on a gene ontology~\cite{yu2018newgoa}, respectively.
% hierarchical labels have been used for genome inference~\cite{valentini2011true,cesa2012synergy} and protein function prediction~\cite{yu2015predicting,yu2018newgoa}.

% \subsection{Memory-Based Models}
% Memory networks~\cite{weston2014memory} are proposed to solve the question and answering problem as an external long-term memory.
% \cite{sukhbaatar2015end} developed a way to train the memory as well as the RNN in an end-to-end fashion.

\section{Targeted Problem and Proposed Model}\label{sec:Genral}
In this section, we first introduce the formulation of many-class few-shot problem in Sec.~\ref{sec:problem}. Then we generally elaborate our network architecture in Sec.~\ref{sec:network}. Details for how to learn a similarity metric for a KNN classifier with attention module and how to update the memory (as a support set of the KNN classifier) are given in Sec.~\ref{sec:attention} and Sec.~\ref{sec:memory} respectively.

\subsection{Problem Formulation}
\label{sec:problem}
We study supervised learning and meta-learning. Given a training set of $n$ samples $\sD = \{(\vx_{i}, y_{i}, z_{i})\}_{i=1}^n$, where each sample $\vx_{i}\in \sX$ is associated with multiple labels. For simplicity, we assume that each training sample is associated with two labels: a fine-class label $y_{i}\in\mathbb Y$ and a coarse-class label $z_i\in \sZ$. The training data is sampled from a data set $\sD$, i.e., $(\vx_{i}, y_{i}, z_{i})\sim\sD$.
Here, $\sX$ denotes the set of samples; $\sY$ denotes the set of all the fine classes, and $\sZ$ denotes the set of all the coarse classes. To define a class hierarchy for $\sY$ and $\sZ$, we further assume that each coarse class $z\in\sZ$ covers a subset of fine classes $\sY_{z}$, and that distinct coarse classes are associated with disjoint subsets of fine classes, i.e., for any $z_1,~z_2\in \sZ$, we have $\sY_{z_1}\cap \sY_{z_2}=\emptyset$. Our goal is fine-class classification by using the class hierarchy information and the coarse labels of the training data. In particular, the supervised learning setting in this case can be formulated as:
\begin{equation}\label{equ:sl}
    \min_\Theta\mathbb E_{(\vx,y,z)\sim\mathcal D}-\log\Pr(y|\vx;\Theta),
\end{equation}
\noindent where $\Theta$ is the model parameters and $\mathbb E$ refers to the expectation w.r.t. the data distribution. In practice, we solve the corresponding empirical risk minimization (ERM) during training, i.e.,
\begin{equation}\label{equ:slerm}
    \min_\Theta\sum_{i=1}^n-\log\Pr(y_i|\vx_i;\Theta).
\end{equation}
In contrast, meta-learning aims to learn a meta-learner model that can be applied to different tasks. Its objective is to maximize the expectation of the prediction likelihood of a task drawn from a distribution of tasks. Specifically, we assume that each task is the classification over a subset of fine classes $T$ sampled from a distribution $\mathcal T$ over all classes, and the problem is formulated as
\begin{equation}\label{equ:ml}
    \min_\Theta\mathbb E_{T\sim\mathcal T}\left[\mathbb E_{(\vx,y,z)\sim\sD_T}-\log\Pr(y|\vx;\Theta)\right],
\end{equation}
where $\sD_T$ refers to the set of samples with label $y_i\in T$. The corresponding ERM is
\begin{equation}\label{equ:mlerm}
    \min_\Theta\sum_{T}\left[\sum_{i\in\mathbb D_{T}}-\log\Pr(y_i|\vx_i;\Theta)\right],
\end{equation}
where $T$ represents a task (defined by a subset of fine classes) sampled from distribution $\mathcal T$, and $\mathbb D_{T}$ is a training set for task $T$ sampled from $\sD_{T}$.

To leverage the coarse class information of $z$, we write $\Pr(y|\vx;\Theta)$ in Eq.~(\ref{equ:sl}) and Eq.~(\ref{equ:ml}) as
\begin{equation}
    \Pr(y|\vx;\Theta)=\sum_{z\in\mathcal Z}\Pr(y|z,\vx;\Theta_f)\Pr(z|\vx;\Theta_c),
\end{equation}
where $\Theta_f$ and $\Theta_c$ are the model parameters for fine classifier and coarse classifier, respectively\footnote{For simplicity, we neglect model parameters $\theta^{CNN}$ for feature extraction here.
}.
Accordingly, given a specific sample $(\vx_i,y_i,z_i)$ with its ground truth labels for coarse and fine classes, we can write $\Pr(y_i|\vx_i;\Theta)$ in Eq.~(\ref{equ:slerm}) and Eq.~(\ref{equ:mlerm}) as follows.
\begin{equation}\label{equ:condprob}
    \Pr(y_i|\vx_i;\Theta)=\Pr(y_i|z_i,\vx_i;\Theta_f)\Pr(z_i|\vx_i;\Theta_c).
\end{equation}
Suppose that a DNN model already produces a logit $a_y$ for each fine class $y$, and a logit $b_z$ for each coarse class $z$, the two probabilities in the right hand side of Eq.~(\ref{equ:condprob}) are computed by applying softmax function to the logit values in the following way.
\begin{align}
\label{equ:softmax}
    \Pr(y_i|z_i,\vx_i;\Theta_f)=\frac{\exp(a_{y_i})}{\sum_{y\in\mathcal Y_{z_i}}\exp(a_{y})},
    \nonumber \\ 
    \Pr(z_i|\vx_i;\Theta_c)=\frac{\exp(b_{z_i})}{\sum_{z\in\mathcal Z}\exp(a_z)}.
\end{align}

Therefore, we integrate multiple labels (both the fine-class label and coarse-class label) in an ERM, whose goal is to maximize the likelihood of the ground truth fine-class label. Given a DNN that produces two vectors of logits $a$ and $b$ for fine class and coarse class respectively, we can train the DNN for supervised learning or meta-learning by solving the ERM problems in Eq.~(\ref{equ:slerm}) or Eq.~(\ref{equ:mlerm}) (with Eq.~(\ref{equ:condprob}) and Eq.~(\ref{equ:softmax}) plugged in).

\subsection{Network Architecture}\label{sec:network}

\begin{figure*}[t!]
\begin{center}
\includegraphics[width=\linewidth]{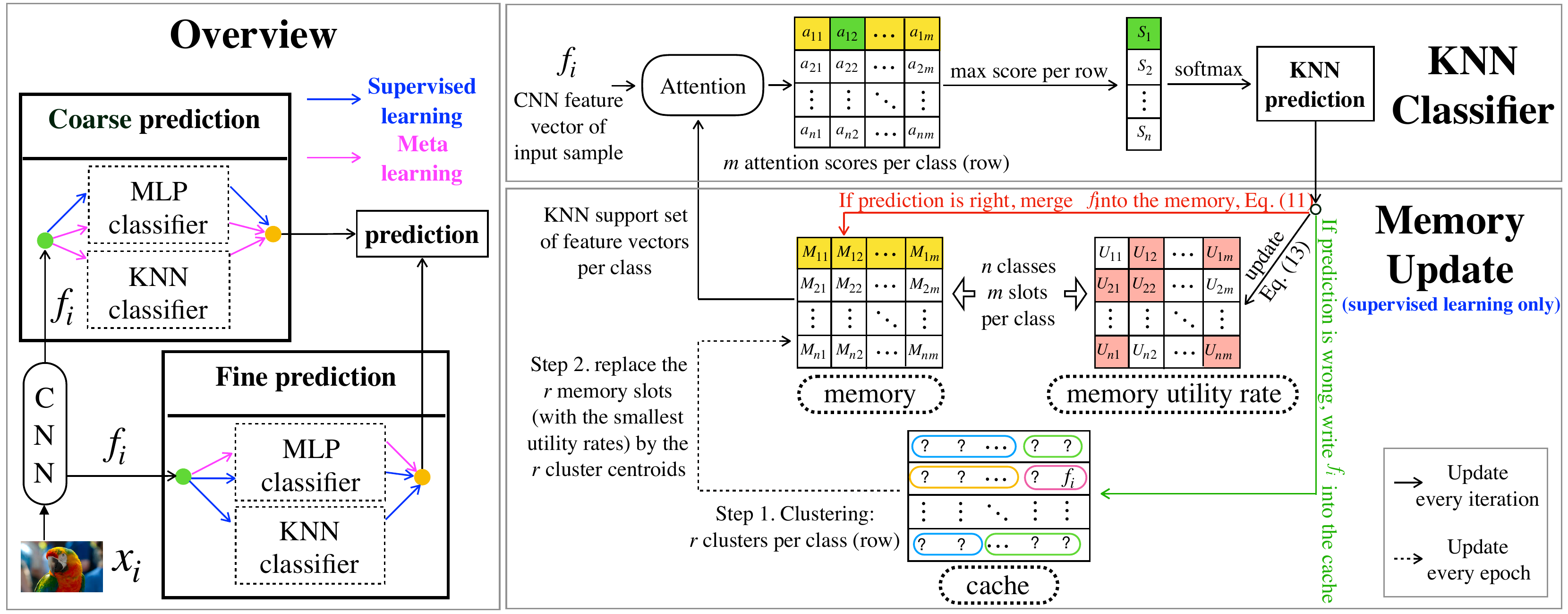}
\end{center}
\caption{
\textbf{Left:} MahiNet. The final fine-class prediction combines predictions based on multiple classes (both fine classes and coarse classes), each of which is produced by an MLP classifier or/and an attention-based KNN classifier.
\textbf{Top right:} KNN classifier with learnable similarity metric and updatable support set.
Attention provides a similarity metric $a_{j,k}$ between each input sample $\vf_{i}$ and a small support set per class stored in memory $\mM_{j,k}$. The learning of KNN classifier aims to optimize 1) the similarity metric parameterized by the attention, detailed in Sec.~\ref{sec:attention}; and 2) a small support set of feature vectors per class stored in memory, detailed in Sec.~\ref{sec:memory}.
\textbf{Bottom right:} The memory update mechanism.
In meta-learning, the memory stores the features of all training samples of a task.
In supervised learning, the memory is updated during training as follows: 
for each sample $\vx_{i}$ within an epoch, if the KNN classifier produces correct prediction, $\vf_{i}$ will be merged into the memory; otherwise, $\vf_{i}$ will be written into a ``cache''.
At the end of each epoch, we apply clustering to the samples per class stored in the cache, and use the resultant centroids to replace $r$ slots of the memory with the smallest utility rate~\cite{gholami2017probabilistic}.}
\label{fig:model}
\end{figure*}

To address MCFS problem in both supervised learning and meta-learning scenarios, we developed a universal model, MahiNet, as in \Figref{fig:model}.
MahiNet uses a CNN to extract features from raw inputs, and then applies two modules to produce coarse-class prediction and fine-class prediction, respectively.
Each module includes one or two classifiers: either an MLP or an attention-based KNN classifier or both.
Intuitively, MLP performs better when data is sufficient, while the KNN classifier is more stable in few-shot scenario. Hence, we always apply MLP to coarse prediction and always apply KNN to fine prediction.
In addition, we use KNN to assist MLP for the coarse module in meta-learning, and use MLP to assist KNN for the fine module in supervised learning.
We develop two mechanisms to make the KNN classifier learnable and be able to quickly adapt to new tasks in the meta-learning scenario.
In the attention-based KNN classifier, an attention module is trained to compute the similarity between two samples, and a re-writable memory is maintained with a highly representative support set for KNN prediction. The memory is updated during training.

Our method for learning a KNN classifier combines the ideas from two popular meta-learning methods, i.e., matching networks~\cite{vinyals2016matching} that aim to learn a similarity metric, and prototypical networks~\cite{snell2017prototypical} that aim to find a representative center per class for NN search. However, our method relies on an augmented memory rather than a bidirectional RNN for retrieving of NN in matching networks. 
In contrast to prototypical networks, which only has one prototype per class, we allow multiple prototypes as long as they can fit in the memory budget. Together these two mechanisms prevent the confusion caused by subtle differences between classes in “many-class” scenario. Notably, MahiNet can also be extended to ``life-long learning'' given this memory updating mechanism. We do not adopt the architecture used in \cite{mishra2018simple} since it requires the representations of all historical data to be stored.

\subsection{Learn a KNN Similarity Metric with an Attention}\label{sec:attention}

In MahiNet, we train an attention module to compute the similarity used in the KNN classifier. 
The attention module learns a distance metric between the feature vector $f_i$ of a given sample $\vx_i$ and any feature vector from the support set stored in the memory. 
Specifically, we use the dot product attention similar to the one adopted in~\cite{vaswani2017attention} for supervised learning, and use an Euclidean distance based attention for meta-learning, following the instruction from~\cite{snell2017prototypical}.
Given a sample $\vx_{i}$, we compute a feature vector $\vf_{i} \in \mathbb{R}^{d}$ by applying a backbone CNN to $\vx_{i}$.
In the memory, we maintain a support set of $m$ feature vectors for each class, i.e., $\mM \in \mathbb{R}^{C \times m \times d}$, where $C$ is the number of classes.
The KNN classifier produces the class probabilities of $\vx_{i}$ by first calculating the attention scores between $\vf_{i}$ and each feature vector in the memory, as follows.
\begin{align}
    a(\vf_{i}, \mM_{j,k}) = \frac{g(\vf_{i}) {\cdot} h(\mM_{j,k})}{\|g(\vf_{i})\|\|h(\mM_{j,k})\|}\nonumber \\
    ~~{\rm or}~~-\|g(\vf_{i})-h(\mM_{j,k})\|_2,~~\forall j\in[C],~k\in[m],
\end{align}
where $g$ and $h$ are learnable transformations for $\vf_{i}$ and the feature vectors in the memory. 

We select the top $K$ nearest neighbors, denoted by $top_{k}$, of $\vf_{i}$ among the $m$ feature vectors for each class $j$, and compute the sum of their similarity scores as the attention score of $\vf_{i}$ to class $j$, i.e.,

% \begin{equation}\label{eq:norm-final-att}
%     s(\vf_{i}, \mM_{j}) = \max_{N\subseteq [m], |N|\leq K} \sum_{k\in N}a(\vf_{i}, \mM_{j,k}),~~\forall j\in[C] .
% \end{equation}
\begin{equation}\label{eq:norm-final-att}
    s(\vf_{i}, \mM_{j}) =  \sum_{k\in [m]} top_{k}( a(\vf_{i}, \mM_{j,k}) ),~~\forall j\in[C].
\end{equation}
We usually find $K=1$ is sufficient in practice. 
The predicted class probability is derived by applying a softmax function to the attention scores of $\vf_{i}$ over all $C$ classes, i.e.,
\begin{align}\label{eq:norm-final-pred}
    \Pr(y_{i}=j)\triangleq\frac{\exp{(s(\vf_{i}, \mM_{j}))}}{\sum_{j'=1}^{C} \exp{(s(\vf_{i}, \mM_{j'}))}},~~\forall j\in[C] .
\end{align}

\subsection{Memory Mechanism for the Support Set of KNN}\label{sec:memory}

Ideally, the memory $\mM \in \mathbb{R}^{C \times m \times d}$ can store all available training samples as the support set of the KNN classifier.
In meta-learning, in each episode, we sample a task with $C$ classes and $m$ training samples per class, and store them in the memory.
Due to the small amount of training data for each task, we can store all of them and thus do not need to update the memory.
In supervised learning, we only focus on one task.
This task usually has a large training set and it is inefficient, unnecessary and too computationally expensive to store all the training set in the memory.
Hence, we set up a budget hyper-parameter $m$ for each class.
$m$ is the maximal number of feature vectors to be stored in the memory for one class.
Moreover, we develop a memory update mechanism to maintain a small memory with diverse and representative feature vectors per class (t-SNE visualization of the diversity and representability of the memory can be found in the experiment section.).
Intuitively, it can choose to forget or merge feature vectors that are no longer representative, and select new important feature vectors into memory.

We will show later in experiments that a small memory can result in sufficient improvement, while the time cost of memory updating is negligible.
Accordingly, the memory in meta-learning does not need to be updated according to a rule, while in supervised learning, we design the writing rule as follows.
During training, for the data that can be correctly predicted by the KNN classifier, we merge its feature with corresponding slots in the memory by computing their convex combination, i.e.,
\begin{equation}\label{eq:memory}
\mM_{j,k} =\{
\begin{array}{ll}
 \gamma \times \mM_{j,k} + (1-\gamma) \times \vf_{i} , & \text{if } \hat{y}_{i} = y_{i} \\
 \mM_{j,k} ,                                   & \text{otherwise}\\
\end{array},
\end{equation}
where $y_{i}$ is the ground truth label, and $\gamma = 0.95$ is a combination weight that works well in most of our empirical studies;
for input feature vector that cannot be correctly predicted, we write it to a cache $\sC = \{\sC_{1},..., \sC_{C}\}$ that stores the candidates written into the memory for the next epoch, i.e.,
\begin{equation}\label{eq:cache}
\sC_{j} =\{
\begin{array}{ll}
\sC_{j},                 & \text{if }  \hat{y}_{i} = y_{i} \\
\sC_{j}\cup \{\vf_{i}\}, & \text{otherwise}\\
\end{array} ,
\end{equation}
Concurrently, we record the utility rate of the feature vectors in the memory, i.e., how many times each feature vector being selected into the $K$ nearest neighbor during the epoch. The rates are stored in a matrix $\mU \in \mathbb{R}^{C \times m}$, and we update it as follows.
\begin{equation}\label{eq:utility}
\mU_{j,k} =\{
\begin{array}{ll}
\mU_{j,k} \times \mu, & \text{if } \hat{y}_{i} = y_{i} \\
\mU_{j,k} \times \eta, & \text{otherwise}\\
\end{array} ,
\end{equation}
where $\mu \in (1,2)$ and $\eta \in (0,1)$ are hyper-parameters.

At the end of each epoch, we apply clustering to the feature vectors per class in the cache, and obtain $r$ cluster centroids as the candidates for memory update in the next epoch. Then, for each class, we replace $r$ feature vectors in the memory that have the smallest utility rate with the $r$ cluster centroids.

\setlength{\textfloatsep}{4.0pt}
\begin{algorithm}[t!]
\caption{Training MahiNet for Supervised Learning}
\label{alg:sup}
\begin{algorithmic}[1]

\REQUIRE Training set $\sD = \{(\vx_{i}, y_{i}, z_{i})\}_{i=1}^n$;\\
\hspace{4mm} Randomly initialized $\theta_{f}^{KNN}$, pre-trained $\theta^{CNN}$, $\theta_{c}^{MLP}$ and $\theta_{f}^{MLP}$; \\
\hspace{4mm} Hyper-parameters: memory update parameters $r$, $\gamma$, $\mu$ and $\eta$; learning rate and its scheduler;
\WHILE{no converge}
\FOR{mini-batch $\{(\vx_{i}, y_{i}, z_{i})\}_{i\in B}$ in $\sD$}
\STATE Compute fine-class logits $a$ and coarse-class logits $b$ from the outputs of MLP/KNN classifiers;
\STATE Apply one step of mini-batch SGD for ERM in Eq.~(\ref{equ:slerm}) (with Eq.~(\ref{equ:condprob}) and Eq.~(\ref{equ:softmax}) plugged in);
\FOR{sample in the mini-batch}
\STATE Update the memory $\mM$ according to Eq.~(\ref{eq:memory});
\STATE Update the utility rate $\mU$ according to Eq.~(\ref{eq:utility});
\STATE Expand the feature cache $\sC$ according to Eq.~(\ref{eq:cache});
\ENDFOR
\ENDFOR

\FOR{each fine class $j$ in $\mathcal Y$}
\STATE Find the indexes of the $r$ smallest values in $\mU_{j}$, denoted as $\{k_{1}, k_{2}, ..., k_{r}\}$;
\STATE Clustering of the feature vectors within cache $\sC_{j}$ to $r$ clusters with centroids $\{c_{1}, c_{2},...,c_{r}\}$;
%\\If $t\in\va$ and $r{\not\in}\va$, then $\mB_{j,t} < \mB_{j,r}$
\STATE Replace the $r$ memory slots by centroids: $\mM_{j,k_{i}} = c_{i}$ for $i\in[r]$;

\ENDFOR
\ENDWHILE
\end{algorithmic}
\end{algorithm}

\section{Training Strategies}

As shown in the network structure in Fig.~\ref{fig:model}, in supervised learning and meta-learning, we use different combinations of MLP and KNN to produce the multi-output predictions of fine-class and coarse-class. The classifiers are combined by summing up their output logits for each class, and a softmax function is applied to the combined logits to generate the class probabilities. Assume the MLP classifiers for the coarse classes and the fine classes are $\phi(\cdot;\theta_c^{MLP})$ and $\phi(\cdot;\theta_f^{MLP})$, the KNN classifiers for the coarse classes and the fine classes are $\phi(\cdot;\theta_c^{KNN})$ and $\phi(\cdot;\theta_f^{KNN})$. In supervised learning, the model parameters are $\theta^{CNN}$, $\Theta_c=\theta_c^{MLP}$ and $\Theta_f=\{\theta_f^{MLP}, \theta_f^{KNN}\}$; in meta-learning setting, the model parameters are $\theta^{CNN}$, $\Theta_c=\{\theta_c^{MLP}, \theta_c^{KNN}\}$ and $\Theta_f=\theta_f^{KNN}$. 

According to Sec.~\ref{sec:problem}, we train MahiNet for supervised learning by solving the ERM problem in Eq.~(\ref{equ:slerm}). For meta-learning, we instead train MahiNet by solving Eq.~(\ref{equ:mlerm}). As previously mentioned, the multi-output logits (for either fine classes or coarse classes) used in those ERM problems are obtained by summing up the logits produced by the corresponding combination of the classifiers.

\subsection{Training MahiNet for Supervised learning}
In supervised learning, the memory update relies heavily on the clustering of the merged feature vectors in the cache. 
To achieve relatively high-quality feature vectors, 
we first pre-train the CNN+MLP model by using standard backpropagation, which minimizes the sum of cross entropy loss on both the coarse-classes and fine-classes. The loss is computed based on the logits of fine-classes and the logits of coarse-classes, which are the outputs of the corresponding MLP classifiers. 
Then, we fint-tune the whole model, which includes the fine-class KNN classifier, with memory update applied. 
Specifically, in every iteration, we firstly update the parameters of the model with mini-batch SGD based on the coarse-level and fine-level logits produced by the MLP and KNN classifiers. Then, the memory that stores a limited number of representative feature maps for every class is updated by the rules explained in Sec.~\ref{sec:memory}, as well as the associated utility rates and feature cache.
The details of the training procedure during the fine-tune stage is explained in Alg.~\ref{alg:sup}.

\begin{algorithm}[t!]
\caption{Training MahiNet for Meta-Learning}
\label{alg:meta}
\begin{algorithmic}[1]
\REQUIRE Training set $\sD = \{(\vx_{i}, y_{i}, z_{i})\}_{i=1}^n$ and fine class set $\sY$;\\
\hspace{4mm} Parameters: randomly initialized $\theta^{CNN}$, $\theta_{c}^{MLP}$, $\theta_{f}^{MLP}$, and $\theta_{f}^{KNN}$; \\
\hspace{4mm} Hyper-parameters: learning rate, scheduler; for each class, number of queries $n_{s}$, support set size $n_{S}$;
\WHILE{not converge}

\STATE Sample a task $T \sim \mathcal T$ as a subset of fine classes $T\subseteq \sY$.
\FOR{class $j$ in $T$}
\STATE Randomly sample $n_s$ data points of class $j$ from $\sD$ to be the support set $\sS_j$ of class $j$.
\STATE Randomly sample $n_q$ data points of class $j$ from $\sD\backslash \sS_j$ to be the query set $\sQ_j$ of class $j$.
\ENDFOR

\FOR{mini-batch from $\sQ$}
\STATE Compute fine-class logits $a$ and coarse-class logits $b$ from the outputs of MLP/KNN classifiers;
\STATE Apply one step of mini-batch SGD for ERM in Eq.~(\ref{equ:mlerm}) (with Eq.~(\ref{equ:condprob}) and Eq.~(\ref{equ:softmax}) plugged in);
\ENDFOR
\ENDWHILE
\end{algorithmic}
\end{algorithm}

\begin{table*}[t!]
\setlength{\tabcolsep}{11pt}
\centering
\caption{
Comparison of the statistics for \textit{mcfs}ImageNet, \textit{mcfs}Omniglot and previously popular datasets for supervised learning and few-shot learning. We propose datasets \textit{mcfs}ImageNet, \textit{mcfs}Omniglot, in which every image is annotated by multiple labels: a coarse label and a fine label. ``\#c'' and ``\#f'' denote the number of coarse classes and fine classes, respectively. ``-'' means ``not applicable''.
}
\begin{threeparttable}
\begin{tabular}{|l|c|c|c|c|c|c|c|c|c|c|c|c|}
\hline
\multirow{3}{*}{}    & \multicolumn{6}{c|}{Meta-Learning} & \multicolumn{4}{c|}{Supervised Learning} & \multirow{3}{*}{\makecell{\#image}} & \multirow{3}{*}{\makecell{image\\size}} \\ \cline{2-11}
                     & \multicolumn{2}{c|}{Train} & \multicolumn{2}{c|}{Val} & \multicolumn{2}{c|}{Test} & \multicolumn{2}{c|}{Train} &    \multicolumn{2}{c|}{Test} & & \\ \cline{2-11}
                     &\multicolumn{1}{c|}{\#c} & \multicolumn{1}{c|}{\#f} &   \multicolumn{1}{c|}{\#c} & \multicolumn{1}{c|}{\#f} &   \multicolumn{1}{c|}{\#c} & \multicolumn{1}{c|}{\#f} &   \multicolumn{1}{c|}{\#c} & \multicolumn{1}{c|}{\#f} &   \multicolumn{1}{c|}{\#c} & \multicolumn{1}{c|}{\#f} & & \\ \hline
             
ImageNet-1k             &  -    &  -       & -    &  -     &  -     &- & 1& 1000 &1 &1000 & 1.43M & 224\\ \hline
\textit{mini}ImageNet   &  1    &  64      & 1    & 16     &  1     & 20& -& -    &- &-    & 0.06M & 84 \\\hline
Omniglot                & 33    & 1028     & 5    &172     &  13  &423&-& -    &- &-    & 0.03M & 28 \\ \hline
\textbf{\textit{mcfs}Omniglot} & 50    & 973     & 50    &244     &  50  &1624&-& -    &- &-    & 0.03M & 28 \\ \hline
\textbf{\textit{mcfs}ImageNet}   & 77   & 482      & 61   & 120    & 68  &152&77 & 754 & 77 & 754 & 0.14M & 112\\ 
\hline
\end{tabular}
\end{threeparttable}
\label{table:dataset}
\end{table*}

\subsection{Training MahiNet for Meta-learning}
When sampling the training/test tasks, we allow unseen fine classes that were not covered in any training task to appear in test tasks, but we fix the ground set of the coarse classes for both training and test tasks. Hence, every coarse class appearing in any test task has been seen during training, but the corresponding fine classes belonging to this coarse class in training and test tasks can be vary. 

In the meta-learning setting, since the tasks in different iterations targets samples from different subsets of classes and thus ideally need different feature map representations~\cite{rusu2018meta}, we do not maintain the memory during  training so every class can have a fixed representation.
Instead, the memory is only maintained within each iteration for a specific task and it stores feature map representations extracted from the support set of the task.
The detailed training procedure can be found in Alg.~\ref{alg:meta}. 
In summary, we generate a training task by randomly sampling a subset of fine classes, and then we randomly sample a support set $\sS$ and a query set $\sQ$ from the associated samples belonging to these classes. We store the CNN feature vectors of $\sS$ in the memory, and train MahiNet to produce correct predictions for the samples in $\sQ$ with the help of coarse label predictions.

\section{Experiments}
In this section, we first introduce two new benchmark datasets specifically designed for MCFS problem, and compare them with other popular datasets in Sec.~\ref{sec:dataset}. Then, we show that MahiNet has better generality and can outperform the specially designed models in both supervised learning (in Sec.~\ref{sec:exp-sup}) and meta-learning (in Sec.~\ref{sec:exp-meta}) scenarios. We also present an ablation study of MahiNet to show the improvements brought by different components and how they interact with each other. In Sec.~\ref{sec:appendix-relation}, we report the comparison with the variants of baseline models in order to show the broad applicability of our strategy of leveraging the hierarchy structures.
In Sec.~\ref{sec:appendix-memory}, we present an analysis of the computational costs caused by the augmented memory in MahiNet.

\subsection{Two New Benchmarks for MCFS Problem: \textit{mcfs}ImageNet \& \textit{mcfs}Omniglot}\label{sec:dataset}
Since existing datasets for both supervised learning and meta-learning settings neither support multiple labels nor fulfill the requirement of ``many-class few-shot'' settings, we propose two novel benchmark datasets specifically for  MCFS Problem, i.e., \textit{mcfs}ImageNet \& \textit{mcfs}Omniglot\footnote{More details about the class hierarchy split information for mcfsImageNet and mcfsOmniglot can be found at: \url{https://github.com/liulu112601/MahiNet}}. The samples in them are annotated by multiple labels of different granularity. We compare them with several existing datasets in Table~\ref{table:dataset}. Our following experimental study focuses on these two datasets.

ImageNet~\cite{imagenet} is one of the most widely used large-scale benchmark dataset for image classification. Although it provides hierarchical information about class labels, it cannot be directly used to test the performance of MCFS learning methods. One main reason is that a fine class in ImageNet may belong to multiple coarse classes, which introduces unnecessary noise in narrowing down the fine-class candidates and is outside the scope of MCFS problem, in which each sample has only one unique coarse-class label. In addition, ImageNet does not satisfy the criteria of ``few-shot'' per class: it has around 1,430 images per class on average.
\textit{mini}ImageNet~\cite{vinyals2016matching} is a widely used benchmark dataset in meta-learning community to test the performance on few-shot learning tasks. \textit{mini}ImageNet is a subset extracted from ImageNet; however, its data are collected from only $80$ fine classes, which is much less than ``many-class'' usually refers to in practice. In addition, it does not provide an associated class hierarchy. 

Hence, to develop a benchmark dataset specifically for the purpose of testing the performance of MCFS learning, we extracted a subset of images from ImageNet and created a dataset called ``\textit{mcfs}ImageNet''.
Table~\ref{table:dataset} compares the statistics of \textit{mcfs}ImageNet with several benchmark datasets.
Comparing to the original ImageNet, we avoided selecting the samples that belong to more than one coarse classes into \textit{mcfs}ImageNet in order to meet the class hierarchy requirements of MCFS problem, i.e., each fine class only belongs to one coarse class. Compared to \textit{mini}ImageNet, \textit{mcfs}ImageNet is about $5\times$ larger, and covers $754$ fine classes in total, which is much more than the $80$ fine classes in \textit{mini}ImageNet. Moreover, on average, each fine class only has $\sim185$ images for training and test, which matches the typical MCFS scenarios in practice. Additionally, the number of coarse classes in \textit{mcfs}ImageNet is $77$, which is much less than $754$ of the fine classes. This is consistent with the data properties found in many practical applications, where the coarse-class labels can only provide weak supervision, but each coarse class has sufficient training samples. Further, we avoided selecting coarse classes which are too broad or contain too many different fine classes. For example, the ``Misc'' class in ImageNet has $20400$ sub-classes, and includes both animal~($3998$ sub-classes) and plant~($4486$ sub-classes). This kind of coarse label covers too many classes and cannot provide valuable information to distinguish different fine classes.

Omniglot~\cite{lake2011one} is a small hand-written character dataset with two-level class labels. However, in their original training/test splitting, the test set contains new coarse classes that are not covered by the training set, since this weakly labeled information is not supposed to be utilized during the training stage. This is inconsistent with the MCFS settings, in which all the coarse classes are exposed in training, but new fine classes can still emerge during test. Therefore, we re-split Omniglot to fulfill the MCFS problem requirement.

\subsection{Supervised Learning Experiments}\label{sec:exp-sup}

\begin{table}
\centering
\setlength{\tabcolsep}{8pt}
\caption{The performance (test accuracy) comparison of different models in the setting of supervised learning on \textit{mcfs}ImageNet.}
\begin{tabular}{lccc}
\toprule
\textbf{Model}            & \textbf{Hierarchy}  & \textbf{\#Params (MB)}  & \textbf{Accuracy} \\ \midrule
{Prototypical Net~\cite{snell2017prototypical}} & N & 11 & 2.7\%  \\
{ResNet18~\cite{He_2016_CVPR}}         & N &  11 & 48.6\% \\ 
\midrule
{MahiNet w/o KNN}  & Y & 11 & 49.1\% \\ 
% \midrule
{MahiNet (ours)}          & Y & 12 &  \textbf{49.9\%} \\ \bottomrule
\end{tabular}
\label{table:exp_sup}
\end{table}

\begin{table*}[t!]
\centering
\setlength{\tabcolsep}{10pt}
\caption{
Test accuracy (\%) of different approaches in meta-learning scenario on \textit{mcfs}ImageNet. Both the average accuracy over 600 test episodes and the corresponding 95\% confidence intervals are reported. In the first row, ``n-k'' represents $n$-way (class) $k$-shot classification task. 
e.g., ``20-10'' refers to ``20-way 10-shot''. To make a fair comparison, all models use the same backbone: ResNet18.
In 50-way experiments, Relation Net stops to improve after the first few iterations and fails to achieve comparable performance (more details in Sec.~\ref{sec:appendix-relation}).
}
\begin{tabular}{lccccccc}
\toprule
\textbf{Model}  &  \textbf{Hierarchy} & \textbf{\#params (MB)} & \textbf{5-10} & \textbf{20-5} & \textbf{20-10} & \textbf{50-5} & \textbf{50-10}  \\
\midrule
\makecell[l]{{ResNet18}\cite{He_2016_CVPR}}
           & N &  11&  60.7 & 58.6 & 67.2 & 48.9 & 56.8 \\
% \midrule
\makecell[l]{{Prototypical Net}\cite{snell2017prototypical}}
                                               & N & 11&  \makecell{78.48$\pm$0.66}& \makecell{67.78$\pm$0.37}& \makecell{70.11$\pm$0.38}& \makecell{57.74$\pm$0.24}& \makecell{62.12$\pm$0.24} \\
% \midrule
\makecell[l]{{Relation Net}\cite{yang2018learning}}
                                               & N & 11&  \makecell{74.12$\pm$0.78}& \makecell{52.66$\pm$0.43}& \makecell{55.45$\pm$0.46}& \makecell{N/A}&
                            \makecell{N/A} \\
% \midrule
\makecell[l]{{MAML}\cite{finn2017model}}
                             & N & 11&  \makecell{61.67$\pm$0.01}& \makecell{47.24$\pm$0.01}& \makecell{48.10$\pm$0.00}&
                             \makecell{11.43$\pm$0.00}&
                             \makecell{11.88$\pm$0.00}
                            \\
% \midrule
\makecell[l]{{Reptile}\cite{nichol2018first}}
                             & N & 11&  \makecell{36.21$\pm$0.01}& \makecell{29.13$\pm$0.01}& \makecell{15.23$\pm$0.01}&
                             \makecell{18.03$\pm$0.00}&
                             \makecell{9.16$\pm$0.00}
                            \\
% \midrule
\makecell[l]{{MahiNet (Ours)}} & Y &  12& 
                                \makecell{\textbf{80.74}$\pm$0.66}&
                                \makecell{\textbf{70.11}$\pm$0.41}&
                                \makecell{\textbf{73.50}$\pm$0.36}& \makecell{\textbf{58.80}$\pm$0.24}&
                              \makecell{\textbf{62.80}$\pm$0.24} \\
\bottomrule
\end{tabular}
\label{table:meta}
\end{table*}

\begin{table*}[t!]
\centering
% \vspace{-0.5em}
\setlength{\tabcolsep}{1.9pt}
\caption{
Ablation studies of MahiNet in the meta-learning setting.  Both the average accuracy over 600 test episodes and the corresponding 95\% confidence intervals are reported. In the first row, ``n-k'' represents $n$-way (class) $k$-shot classification task. ``Mem-1'' stores  the  average  feature  of  all  training  samples  for  each class in one task; ``Mem-2'' stores all features of the training samples  in  one  task.
}
\begin{tabular}{lllllllllll}
\toprule
\multicolumn{2}{c}{\textbf{Memory}}   & \multicolumn{1}{c}{\multirow{2}{*}{\textbf{Attention}}} &  \multicolumn{1}{c}{\multirow{2}{*}{\textbf{Hierarchy}}}   & \multicolumn{1}{c}{\multirow{2}{*}{\textbf{MLP-classifier}}} & \multicolumn{1}{c}{\multirow{2}{*}{\textbf{KNN-classifier}}} &\multicolumn{1}{c}{\multirow{2}{*}{\textbf{5-10}}} &\multicolumn{1}{c}{\multirow{2}{*}{\textbf{20-5}}} &\multicolumn{1}{c}{\multirow{2}{*}{\textbf{20-10}}} &\multicolumn{1}{c}{\multirow{2}{*}{\textbf{50-5}}}&\multicolumn{1}{c}{\multirow{2}{*}{\textbf{50-10}}} \\ 
\multicolumn{1}{c|}{Mem-1} & 
\multicolumn{1}{c}{Mem-2} & \multicolumn{1}{c}{}
  & \multicolumn{1}{c}{\textbf{}} &  \multicolumn{1}{c}{\textbf{}} &  \multicolumn{1}{c}{} &  \multicolumn{1}{c}{} &  \multicolumn{1}{c}{} &  \multicolumn{1}{c}{} &  \multicolumn{1}{c}{} \\ \midrule
        \multicolumn{1}{c}{$\checkmark$} &  & &  &  \multicolumn{1}{c}{$\checkmark$} &        \multicolumn{1}{c}{$\checkmark$}   &  \makecell{79.04$\pm$0.67}  &  \makecell{68.46$\pm$0.38}  &  \makecell{71.13$\pm$0.38} &  \makecell{58.09$\pm$0.24} & \makecell{62.18$\pm$0.22} \\
 &     \multicolumn{1}{c}{$\checkmark$} &  &  & \multicolumn{1}{c}{$\checkmark$} &    \multicolumn{1}{c}{$\checkmark$}   &  \makecell{77.41$\pm$0.71} & \makecell{66.89$\pm$0.40}&    \makecell{71.72$\pm$0.37}&   \makecell{55.25$\pm$0.23} & \makecell{59.38$\pm$0.23} \\
  &     \multicolumn{1}{c}{$\checkmark$} & \multicolumn{1}{c}{$\checkmark$}  & &  \multicolumn{1}{c}{$\checkmark$} & \multicolumn{1}{c}{$\checkmark$}   &  \makecell{76.85$\pm$0.67}&       \makecell{66.43$\pm$0.41}& \makecell{70.01$\pm$0.38}& \makecell{55.13$\pm$0.23} & \makecell{59.22$\pm$0.23}  \\
 \multicolumn{1}{c}{$\checkmark$}  &     \multicolumn{1}{c}{$\checkmark$} & \multicolumn{1}{c}{$\checkmark$}  & &  \multicolumn{1}{c}{$\checkmark$} & \multicolumn{1}{c}{$\checkmark$}   &  \makecell{78.27$\pm$0.68}&
\makecell{67.03$\pm$0.41}& \makecell{71.20$\pm$0.37}& \makecell{57.98$\pm$0.24} & \makecell{62.40$\pm$0.23} \\
 \multicolumn{1}{c}{$\checkmark$}  &     \multicolumn{1}{c}{$\checkmark$} & \multicolumn{1}{c}{$\checkmark$}  & \multicolumn{1}{c}{$\checkmark$}  & & \multicolumn{1}{c}{$\checkmark$}   & \makecell{79.61$\pm$0.67} & \makecell{69.49$\pm$0.40} & \makecell{72.52$\pm$0.37} & \makecell{58.48$\pm$0.24} &  \makecell{62.62$\pm$0.24} \\ 
  \multicolumn{1}{c}{$\checkmark$}  &     \multicolumn{1}{c}{$\checkmark$} & \multicolumn{1}{c}{$\checkmark$}  & \multicolumn{1}{c}{$\checkmark$}  & \multicolumn{1}{c}{$\checkmark$} & & \makecell{79.06$\pm$0.66} & \makecell{69.10$\pm$0.41} & \makecell{72.15$\pm$0.36} & \makecell{58.30$\pm$0.23} & \makecell{62.51$\pm$0.24} \\       
\multicolumn{1}{c}{$\checkmark$} &     &\multicolumn{1}{c}{$\checkmark$}  & \multicolumn{1}{c}{$\checkmark$} &  \multicolumn{1}{c}{$\checkmark$} &        \multicolumn{1}{c}{$\checkmark$}   &   \makecell{80.64$\pm$0.64}   &  \makecell{68.99$\pm$0.40}  &  \makecell{72.78$\pm$0.37}  &  \makecell{58.56$\pm$0.25} & \makecell{62.70$\pm$0.24} \\
 \multicolumn{1}{c}{$\checkmark$}  &     \multicolumn{1}{c}{$\checkmark$} & \multicolumn{1}{c}{$\checkmark$}  & \multicolumn{1}{c}{$\checkmark$}  &  \multicolumn{1}{c}{$\checkmark$} & \multicolumn{1}{c}{$\checkmark$}   &   \makecell{\textbf{80.74}$\pm$0.66}& \makecell{\textbf{70.11}$\pm$0.41}& \makecell{\textbf{73.50}$\pm$0.36}& \makecell{\textbf{58.80}$\pm$0.24} & \makecell{\textbf{62.80}$\pm$0.24} \\
\bottomrule
\end{tabular}
\label{table:case-study}
% \vspace{-1.5em}
\label{table:meta-ablation}
\end{table*}

\subsubsection{Setup}
We use ResNet18 \cite{He_2016_CVPR} as the backbone CNN. 
The transformation functions $g$ and $h$ in the attention module are two fully connected layers followed by group normalization~\cite{GroupNorm2018} with a residual connection.
% This feature serves as the features of the MLP fine classifier. 
% We map this feature to the coarse MLP classifier domain by a fully connected layer. 
% We map the feature to the fine KNN classifier domain by two convolutional layers with a residual connection. 
% We use one fully connected layer for the transformation functions $g$ and $h$, respectively.
% % $\mu$ is 1.05, $\eta$ is 0.95
We set the memory size to $m=12$ and the number of clusters to $r=3$, which can achieve a better trade-off between the memory cost and performance. 
Batch normalization \cite{ioffe2015batch} is applied after each convolution and before activation.
During pre-training, we apply the cross entropy loss on the probability predictions in Eq.~(\ref{equ:softmax}).
During fine-tuning, we fix the $\theta^{CNN}$, $\theta^{MLP}_{c}$, and $\theta_{f}^{MLP}$ to ensure the fine-tuning process is stable.
We use SGD with a mini-batch size of $128$ and a cosine learning rate scheduler with an initial learning rate $0.1$.
$\mu=1.05$, $\eta=0.95$, a weight decay of $0.0001$, and a momentum of $0.9$ are used.
We train the model for $100$ epochs during pre-training and $90$ epochs for the fine-tuning.
All hyperparameters are tuned on 20\% samples randomly sampled from the training set.
we use 20\% data randomly sampled from the training set to serve as the validation set to tune all the hyperparameters.

\subsubsection{Experiments on \textit{mcfs}ImageNet}
Table~\ref{table:exp_sup} compares MahiNet with the supervised learning model (i.e., ResNet18) and meta-learning model (i.e., prototypical networks) in the setting of supervised learning.
The results show that MahiNet outperforms the specialized models, such as ResNet18 in MCFS scenario. 

We also show the number of parameters for every method, which indicates that our model only has 9\% more parameters than the baselines. Even using the version without KNN (which has the same number of parameters as baselines) can still outperform them.

% Prototypical Net is a meta-learning model designed to solve few-shot classification problems.
% We train it in a supervised learning manner (i.e., on a single task with many classes and relatively much more samples per class), and include it in the comparison to test its performance on MCFS problem.
% Prototypical network fails to solve MCFS problem in the supervised learning scenario.

Prototypical networks have been specifically designed for few-shot learning. Although it can achieve promising performance on few-shot tasks each covering only a few classes and samples, it fails when directly used to classify many classes in the classical supervised learning setting, as shown in Table~\ref{table:exp_sup}. A primary reason, as we discussed in Section 1, is that the resulted prototype suffers from high variance (since it is simply the average over an extremely few amount of samples per class) and cannot distinguish each class from tens of thousands of other classes (few-shot learning task only needs to distinguish it from 5-10 classes). Hence, dedicated modifications is necessary to make prototypical networks also work in the supervised learning setting. One motivation of this paper is to develop a model that can be trained and directly used in both settings.

The failure of Prototypical networks in supervised learning setting is not too surprising since for prototypical network the training and test task in the supervised setting is significantly different. We keep its training stage the same (i.e., target a task of 5 classes in every episode). In the test stage, we use the prototypes as the per-class means of training samples in each class to perform classification over all available classes on the test set data. Note the number of classes in the training tasks and the test task are significantly different, which introduces a large gap between training and test that leads to the failure. This can also be validated by observing the change on performance when we intentionally reduce the gap: the accuracy improves as the number of classes in each training task increases.

Another potential method to reduce the training-test gap is to increase the number of classes/ways per task to the same number of classes in supervised learning. However, prototypical net requires an prohibitive memory cost in such many class setting, i.e, it needs to keep 754(number of classes)$\times$2(1 for support set, 1 for query set)=1512 samples in memory for 1 shot learning setting and 754$\times$6(5 for support set, 1 for query set) = 4536 samples for 5 shots setting. Unlike classical supervised learning models which only needs to keep the model parameters in memory, prototypical net needs to build prototypes on the fly from the few-shot samples in the current episode, so at least one sample should be available to build the prototypes and at least one sample for query.

To separately measure the contribution of the class hierarchy and the attention-based KNN classifier, we conduct an ablation study that removes the KNN classifier from MahiNet.
The results show that MahiNet still outperforms ResNet18 even when only using the extra coarse-label information and an MLP classifier for fine classes during training. 
Using a KNN classifier further improves the performance since KNN is more robust in the few-shot fine-level classifications. In supervised learning scenario, memory update is applied in an efficient way. 
1) For each epoch, the average clustering time is 30s and is only $7.6\%$ of the total epoch time (393s). 
2) Within an epoch, the memory update time (0.02s) is only $9\%$ of the total iteration time (0.22s).

\subsection{Meta-Learning Experiments}\label{sec:exp-meta}

\subsubsection{Setup}
We use the same backbone CNN (i.e., ResNet18) and transformation functions (i.e., $g$, and $h$) for attention as in supervised learning.
In each task, we sample the same number of classes for training and test, and follow the training procedure in \cite{snell2017prototypical}. For a fair comparison, all baselines and our model are trained on the same number of classes for all training tasks, since increasing the number of classes in training tasks improves the performance as indicated in~\cite{snell2017prototypical}.
We initialize the learning rate as $10^{-3}$ and halve it after every 10k iterations. Our model is trained by ADAM~\cite{kingma2015adam} using a mini-batch size of $128$ (only in the pre-training stage), a weight decay of $0.0001$, and a momentum of $0.9$. We train the model for 25k iterations in total. Given the class hierarchy, the training objective sums up the cross entropy losses computed over the coarse class and over the fine classes, respectively.
All hyper-parameters are tuned based on the validation set introduced in Table~\ref{table:dataset}. 
Every baseline and our method use the same training set and test test, and the test set is inaccessible during training so any data leakage is prohibited. In every experimental setting, we make the comparison fair by applying the same backbone, same training and test data, the same task setting (we follow the most popular N way K shot problem setting) and the same training strategy (including how to sample tasks and the value of N and K for meta-learning, since higher N and K generally lead to better performance as indicated in~\cite{snell2017prototypical}).

\subsubsection{Experiments on \textit{mcfs}ImageNet}
Table~\ref{table:meta} shows that MahiNet outperforms the supervised learning baseline (ResNet18) and the meta-learning baseline (Prototypical Net).
For ResNet18, we use the fine-tuning introduced in~\cite{finn2017model}, in which the trained network is first fine-tuned on the training set (i.e., the support set) provided in the new task in the test stage and then tested on the associated query set.
To evaluate the contributions of each component in MahiNet, we show results of several variants in Table~\ref{table:meta-ablation}.
``Attention'' refers to the parametric functions for $g$ and $h$, otherwise we use identity mapping.
``Hierarchy'' refers to the assist of class hierarchy.
For a specific task, ``Mem-1'' stores the average embedding over all training samples for each class in one task; ``Mem-2'' stores all the embeddings of the training samples in one task; ``Mem-3'' is the union of  ``Mem-1'' and ``Mem-2''.
Table~\ref{table:meta-ablation} implies: (1) Class hierarchy information can incur steady performance across all tasks; (2) Combining ``Mem-1'' and ``Mem-2'' outperforms using either of them independently; 
(3) Attention brings significant improvement to MCFS problem only when being trained with class hierarchy. Because the data is usually insufficient to train a reliable similarity metric to distinguish all fine classes, but distinguishing a few fine classes in each coarse class is much easier.
The attention module is merely parameterized by the two linear layers $g$ and $h$, which are usually not expressive enough to learn the complex metric without the side information of class hierarchy.
With the class hierarchy as a prior knowledge, the attention module only needs to distinguish much fewer fine classes within each coarse class, and the learned attention can faithfully reflect the local similarity within each coarse class.
We also show the number of parameters for every method in the table. Our method only has 9\% more parameters compared to baselines.

\subsubsection{Experiments on \textit{mcfs}Omniglot}
We conduct experiments on the secondary benchmark \textit{mcfs}Omniglot. We use the same training setting as for \textit{mcfs}ImageNet.
Following \cite{meta-mem-aug}, \textit{mcfs}Omniglot is augmented with rotations by multiples of 90 degrees.
% We do not compare with ResNet18 on \textit{mcfs}Omniglot, since \textit{mcfs}Omniglot is a $28\times28$ small dataset, which would be easy for ResNet18 to get overfitting.
In order to make an apple-to-apple comparison to other baselines, we follow the same backbone used in their experiments.
Therefore, we use four consecutive convolutional layers, each followed by a batch normalization layer, as the backbone CNN and compare MahiNet with prototypical networks as in Table~\ref{table:meta-omniglot}.
We do ablation study on MahiNet with/without hierarchy and MahiNet with different kinds of memory. ``Mem-3'', i.e., the union of ``Mem-1'' and ``Mem-2'', outperforms ``Mem-1'', and ``Attention'' mechanism can improve the performance.
Additionally, MahiNet outperforms other compared methods, which indicates the class hierarchy assists to make more accurate predictions.
In summary, experiments on the small-scale and large-scale datasets show that class hierarchy brings a stable improvement.
\begin{table*}[t!]
\centering
\setlength{\tabcolsep}{11pt}
\caption{
Test accuracy (\%) of basedline, our method and its variants in the meta-learning scenario on \textit{mcfs}Omniglot. Both the average accuracy over 600 test episodes and the corresponding 95\% confidence intervals are reported. In the first row, ``n-k'' represents $n$-way (class) $k$-shot classification task.
e.g., ``20-10'' refers to ``20-way 10-shot''.
}
\begin{tabular}{lcccc}
\toprule
\textbf{Model}                               & \textbf{Hierarchy} & \textbf{5-5} & \textbf{20-5} &  \textbf{50-5}   \\
\midrule
\makecell[l]{{Reptile}~\cite{nichol2018first}}
                                               & N & \makecell{54.70$\pm$0.02}& \makecell{17.10$\pm$0.01}& 
                            \makecell{8.34$\pm$0.02} \\
\makecell[l]{{MAML}~\cite{finn2017model}}
                                               & N & \makecell{74.60$\pm$0.01}& \makecell{21.61$\pm$0.00}& 
                                        \makecell{9.27$\pm$0.20} \\
\makecell[l]{{Prototypical Net}~\cite{snell2017prototypical}}
                                               & N & \makecell{99.10$\pm$0.15}& \makecell{98.84$\pm$0.11}& 
                                        \makecell{97.94$\pm$0.08} \\
\midrule
\makecell[l]{{MahiNet (Mem-1)}}
                                             & N & \makecell{99.17$\pm$0.16}& \makecell{98.82$\pm$0.11}& \makecell{97.96$\pm$0.09}\\ 
\makecell[l]{{MahiNet (Mem-3)}}
                                             & N & \makecell{99.31$\pm$0.18}& \makecell{98.89$\pm$0.11}& \makecell{97.93$\pm$0.09}\\
% \midrule
\makecell[l]{{MahiNet} (Mem-3)} & Y &
                                \makecell{\textbf{99.40}$\pm$0.15}& \makecell{\textbf{99.00}$\pm$0.16}&
                                \makecell{\textbf{98.10}$\pm$0.17} \\
\bottomrule
\end{tabular}
\label{table:meta-omniglot}
\end{table*}

\subsection{Comparison to Variants of Relation Net}
\label{sec:appendix-relation}

\subsubsection{Relation network with class hierarchy} In order to demonstrate that the class hierarchy information and the multi-output model not only improves MahiNet but also other existing models, we train relation network with class hierarchy in the similar manner as we train MahiNet. The results are shown in Table ~\ref{table:rela-appendix}. It indicates that the class hierarchy also improves the accuracy of relation network by more than $1\%$, which verifies the advantage and generality of using class hierarchy in other models. Although the relation net with class hierarchy achieves a better performance, MahiNet still outperforms this improved variant due to its advantage in its model architecture, which are specially designed for utilizing the class hierarchy in the many-class few-shot scenario.

\begin{table}[ht!]
\centering
\setlength{\tabcolsep}{8pt}
\caption{
The improvement of class hierarchy on relation network on \textit{mcfs}ImageNet. The average test accuracy over 600 test episodes and the corresponding 95\% confidence intervals are reported.
}
\resizebox{\linewidth}{!}{
\begin{tabular}{lcccc}
\toprule
\textbf{Model}                               & \textbf{Hierarchy} & \textbf{5 way 5 shot} & \textbf{5 way 10 shot}  \\
\midrule

\makecell[l]{{Relation Net}\cite{yang2018learning}}
                      & N &  \makecell{63.02$\pm$0.87}  & \makecell{74.12$\pm$0.78} \\
\makecell[l]{{Relation Net}\\}
                       & Y  & \makecell{66.82$\pm$0.86} & \makecell{75.31$\pm$0.90}\\
\midrule
\makecell[l]{{MahiNet (Mem-{3})}} & Y & 
                                \makecell{\textbf{74.98}$\pm$0.75} & \makecell{\textbf{80.74}$\pm$0.66}  \\
\bottomrule
\end{tabular}}
\label{table:rela-appendix}
\end{table}

\subsubsection{Relation Network in many-class setting} When relation network is trained in many-class settings using the same training strategy from~\cite{yang2018learning}, we found that the network usually stops to improve and stays near a sub-optimal solution.
Specifically, after the first few iterations, the training loss still stays at a high level and the training accuracy still stays low. 
They do not change too much even after hundreds of iterations, and this problem cannot be solved after trying different initial learning rates. 
% We show the training loss and training accuracy for the first 100 iterations under different learning rate as Fig.~\ref{fig:relation-loss}. 
The training loss quickly converges to around 0.02 and the training accuracy stays at $\sim2\%$ no matter what learning rate is applied to the training procedure. 
Relation net on other low way settings have a competitive performance as shown in Table~\ref{table:rela-appendix}, which is also the settings reported in their papers.
We use the same hyper-parameters as their paper for both few-class and many-class settings and the difference is only the parameter of the number of way. 
Even if relation network performs well in the few-class settings, its performance dramatically degrade when given the challenging high-way settings.
One potential reason for this is that relation net uses MSE loss to regress the relation
score to the ground truth: matched pairs have similarity
1 and mismatched pair have similarity 0.
When the number of ways increases, the imbalance between the number of matched pairs and unmatched pairs grows and training is mainly to optimize the loss generated by the mismatched pairs instead of the matched ones, so that the objective focuses more on how to avoid mismatching instead of matching, i.e., predicting correctly, and makes the accuracy more difficult to get improvement. 
This suggests that relation networks may need more tricky way to train it successfully in a many-class setting compared to the few-class settings and traditional few-class few-shot learning approaches may not be directly applied in many-class few-shot problems.
This is also one of the drawbacks for relation net and how to make the model more robust to hyper-parameter tuning is beyond the scope of our paper.

% \begin{figure}[ht]
% \begin{center}
% \includegraphics[width=\linewidth]{resources/iter-loss-acc.pdf}
% \end{center}
% \caption{
% The many-class few-shot problem is challenging for Relation Network. The training loss and accuracy for 50 way 5 shot relation net during the first 100 iterations under different learning rate. Whatever the learning rate is, it quickly converges to a sub-optimal point: The training loss: $\approx$0.02 and the training accuracy: $\approx$2\%.
% }
% \label{fig:relation-loss}
% \end{figure}

\subsection{Analysis of Hierarchy}

The main reasons of the improvement in our paper are: (1) our model can utilize prior knowledge from the class hierarchy; (2) we use an effective memory update scheme to iteratively and adaptively refine the support set. We show how much improvement the class hierarchy can bring to our model in Table~\ref{table:meta-ablation} and to other baselines in Table～\ref{table:rela-appendix}. The upper half of Table~\ref{table:meta-ablation} are variants of our model without using hierarchy while the lower half are variants with hierarchy. It shows that the variants with hierarchy consistently outperform the ones without using hierarchy. To see whether hierarchy can help to improve other baselines, in Table～\ref{table:rela-appendix}, we add hierarchy information to relation net (while keeping the other components the same) and it achieves 1-4\% improvement on accuracy. 

% The main source of the improvement in our paper comes from the hierarchy. Following our motivation that hierarchy can alleviate the challenge of many-class via separating them into small groups, we anaylyze how hierarchy can improve the performance on our model in Table~\ref{table:meta-ablation} and on other baselines in Table \ref{table:rela-appendix}. The upper half of Table 5 are variants of our model without hierarchy and the lower half are variants with hierarchy. In general, variants with hierarchy show more competitive performance compared to those without. In order to figure out if hierarchy can help to improve other baselines, as shown in Table \ref{table:rela-appendix}, we add hierarchy information to relation net and it shows 1-4\% accuracy improvement while keeping the other components the same.

\subsection{Analysis of Augmented Memory}
\label{sec:appendix-memory}

\subsubsection{Visualization}
\begin{figure}[ht]
\begin{center}
\includegraphics[width=\linewidth]{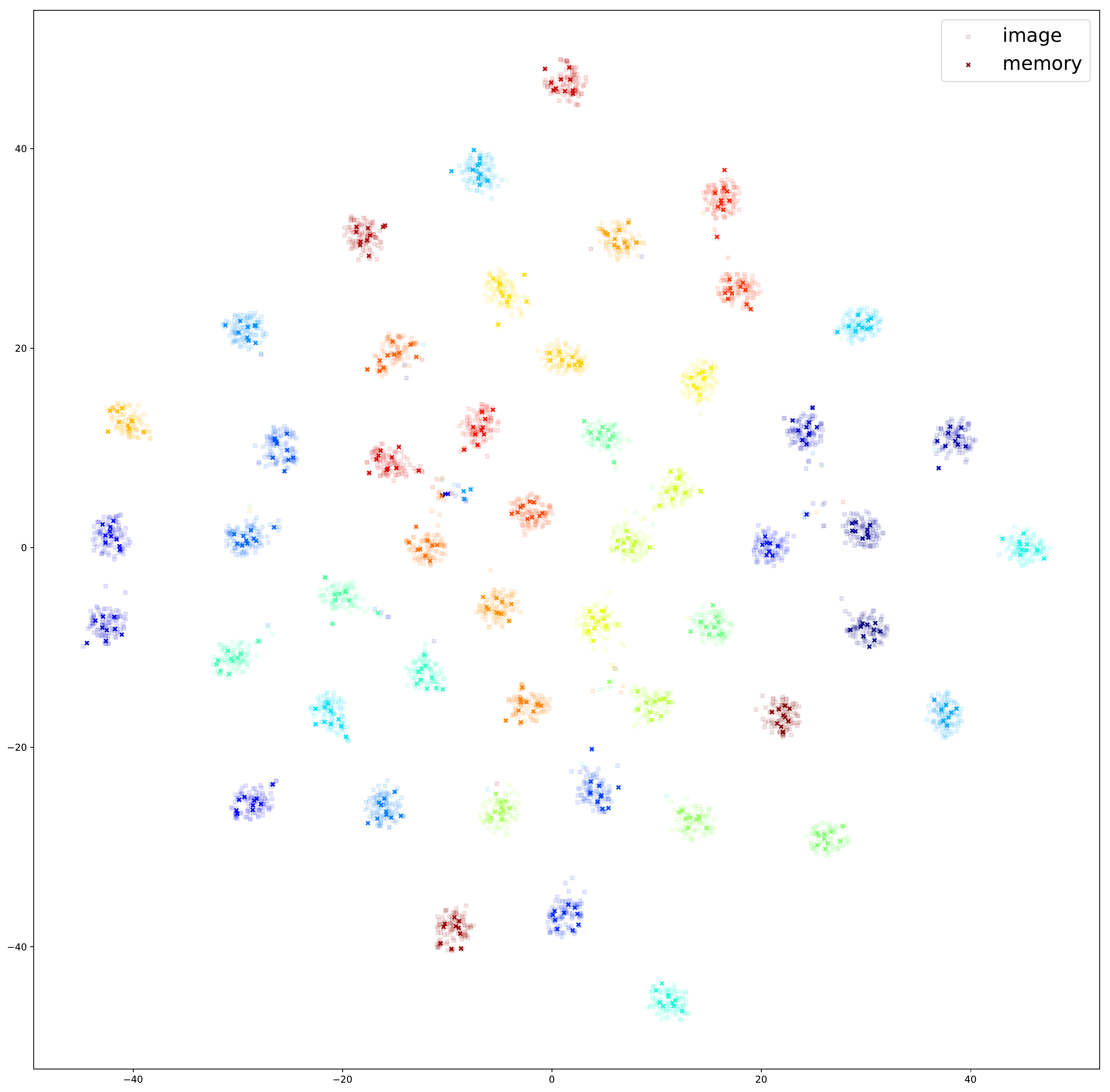}
\end{center}
\caption{The t-SNE visualization for memory maintained in the setting of supervised learning. We randomly sample 50 out of 754 fine classes shown as different colors. The sampled image feature of the training samples and the stored memory feature for one class share the same color while the image feature has higher transparency. For each class, the image features surround the memory feature and features from different classes are dispersed from each other for classification.}
\label{fig:t-sne}
\end{figure}

In order to show how representative and diverse the feature vectors selected into the memory slots are, we visualize them together with the unselected images' feature vectors using t-SNE in Fig.~\ref{fig:t-sne}\cite{maaten2008visualizing}. In particular, we randomly sample 50 fine classes marked by different colors. Within every class, we show both the feature vectors selected into the memory and the feature vectors of the original images from the same class. The features in the memory (crosses with a lower transparency) can cover most areas where the image features (dots of a higher transparency) are located. It implies that the  highly selected feature vectors in memory are diverse and sufficiently representative of the whole class.

\subsubsection{Memory Cost} In the supervised learning experiments, the memory load required by MahiNet is only $754\times 12/125,321 = 7.2\%$ ($12$ samples per class for all the $754$ fine classes, while the training set includes $125,321$ images in total) of the space needed to store the whole training set. We tried to increase the memory size to about $10\%$ of the training set, but the resultant improvement on performance is negligible compared to the extra computation costs, meaning that increasing the memory costs is unnecessary for MahiNet. In contrast, for meta-learning, in each task, every class only has few-shot samples, so the memory required to store all the samples is very small. For example, in the 20-way 1-shot setting, we only need to store $20$ feature vectors in the memory. Therefore, our memory costs in both supervised learning and meta-learning scenarios can be kept small and the proposed method is memory-efficient.

\section{Conclusion}
In this paper, we study a new problem ``many-class few-shot''(MCFS), which is a more challenging problem commonly encountered in practice compared to the previously studied ``many-class many-shot'', ``few-class few-shot'' and ``few-class many-shot'' problems. 
We address the MCFS problem by training a multi-output model producing coarse-to-fine predictions, which leverages the class hierarchy information and explore the relationship between fine and coarse classes.
We propose ``Memory-Augmented Hierarchical-Classification Network (MahiNet)'', which integrates both MLP and learnable KNN classifiers with attention modules to produce reliable prediction of both the coarse and fine classes.
In addition, we propose two new benchmark datasets with a class hierarchy structure and multi-label annotations for the MCFS problem, and show that MahiNet outperforms existing methods in both supervised learning and meta-learning settings on the benchmark datasets without losing advantages in efficiency.

\bibliographystyle{IEEEtran}
\bibliography{IEEEabrv,IEEEtran}

% Generated by IEEEtran.bst, version: 1.12 (2007/01/11)
\begin{thebibliography}{10}
\providecommand{\url}[1]{#1}
\csname url@samestyle\endcsname
\providecommand{\newblock}{\relax}
\providecommand{\bibinfo}[2]{#2}
\providecommand{\BIBentrySTDinterwordspacing}{\spaceskip=0pt\relax}
\providecommand{\BIBentryALTinterwordstretchfactor}{4}
\providecommand{\BIBentryALTinterwordspacing}{\spaceskip=\fontdimen2\font plus
\BIBentryALTinterwordstretchfactor\fontdimen3\font minus
  \fontdimen4\font\relax}
\providecommand{\BIBforeignlanguage}[2]{{%
\expandafter\ifx\csname l@#1\endcsname\relax
\typeout{** WARNING: IEEEtran.bst: No hyphenation pattern has been}%
\typeout{** loaded for the language `#1'. Using the pattern for}%
\typeout{** the default language instead.}%
\else
\language=\csname l@#1\endcsname
\fi
#2}}
\providecommand{\BIBdecl}{\relax}
\BIBdecl

\bibitem{He_2016_CVPR}
K.~He, X.~Zhang, S.~Ren, and J.~Sun, ``Deep residual learning for image
  recognition,'' in \emph{Proc. {IEEE} Conf. Comput. Vis. Pattern Recognit.},
  2016, pp. 770--778.

\bibitem{huang2017densely}
G.~Huang, Z.~Liu, L.~Van Der~Maaten, and K.~Q. Weinberger, ``Densely connected
  convolutional networks,'' in \emph{Proc. {IEEE} Conf. Comput. Vis. Pattern
  Recognit.}, 2017, pp. 4700--4708.

\bibitem{metaRNN}
M.~Andrychowicz, M.~Denil, S.~G\'{o}mez, M.~W. Hoffman, D.~Pfau, T.~Schaul,
  B.~Shillingford, and N.~de~Freitas, ``Learning to learn by gradient descent
  by gradient descent,'' in \emph{Proc. Advances Neural Inf. Process. Syst.},
  2016, pp. 3981--3989.

\bibitem{learn2opt}
K.~Li and J.~Malik, ``Learning to optimize,'' in \emph{Proc. Int. Conf. Learn.
  Representations}, 2017.

\bibitem{ravi2017optimization}
S.~Ravi and H.~Larochelle, ``Optimization as a model for few-shot learning,''
  in \emph{Proc. Int. Conf. Learn. Representations}, 2017.

\bibitem{finn2017model}
C.~Finn, P.~Abbeel, and S.~Levine, ``Model-agnostic meta-learning for fast
  adaptation of deep networks,'' in \emph{Proc. Int. Conf. Machine Learning},
  2017, pp. 1126--1135.

\bibitem{vinyals2016matching}
O.~Vinyals, C.~Blundell, T.~Lillicrap, D.~Wierstra \emph{et~al.}, ``Matching
  networks for one shot learning,'' in \emph{Proc. Advances Neural Inf.
  Process. Syst.}, 2016, pp. 3630--3638.

\bibitem{snell2017prototypical}
J.~Snell, K.~Swersky, and R.~Zemel, ``Prototypical networks for few-shot
  learning,'' in \emph{Proc. Advances Neural Inf. Process. Syst.}, 2017, pp.
  4077--4087.

\bibitem{douze2018low}
M.~Douze, A.~Szlam, B.~Hariharan, and H.~J{\'e}gou, ``Low-shot learning with
  large-scale diffusion,'' in \emph{Proc. {IEEE} Conf. Comput. Vis. Pattern
  Recognit.}, 2018, pp. 3349--3358.

\bibitem{imagenet}
J.~Deng, W.~Dong, R.~Socher, L.-J. Li, K.~Li, and L.~Fei-Fei, ``{ImageNet}: A
  large-scale hierarchical image database,'' in \emph{Proc. {IEEE} Conf.
  Comput. Vis. Pattern Recognit.}, 2009, pp. 248--255.

\bibitem{lake2011one}
B.~Lake, R.~Salakhutdinov, J.~Gross, and J.~Tenenbaum, ``One shot learning of
  simple visual concepts,'' in \emph{Proc. of the Annual Meeting of the
  Cognitive Science Society (CogSci)}, 2011, pp. 2568--2573.

\bibitem{yang2018learning}
F.~S.~Y. Yang, L.~Zhang, T.~Xiang, P.~H. Torr, and T.~M. Hospedales, ``Learning
  to compare: Relation network for few-shot learning,'' in \emph{Proc. {IEEE}
  Conf. Comput. Vis. Pattern Recognit.}, 2018, pp. 1199--1208.

\bibitem{fei2006one}
L.~Fei-Fei, R.~Fergus, and P.~Perona, ``One-shot learning of object
  categories,'' \emph{{IEEE} Trans. Pattern Anal. Mach. Intell.}, vol.~28,
  no.~4, pp. 594--611, 2006.

\bibitem{wong2015one}
A.~Wong and A.~L. Yuille, ``One shot learning via compositions of meaningful
  patches,'' in \emph{Proc. {IEEE} Int. Conf. Comput. Vis.}, 2015, pp.
  1197--1205.

\bibitem{lake2013one}
B.~M. Lake, R.~R. Salakhutdinov, and J.~Tenenbaum, ``One-shot learning by
  inverting a compositional causal process,'' in \emph{Proc. Advances Neural
  Inf. Process. Syst.}, 2013, pp. 2526--2534.

\bibitem{Wang-2018-105271}
Y.~Wang, R.~Girshick, M.~Hebert, and B.~Hariharan, ``Low-shot learning from
  imaginary data,'' in \emph{Proc. {IEEE} Conf. Comput. Vis. Pattern
  Recognit.}, 2018, pp. 7278--7286.

\bibitem{naik1992meta}
D.~K. Naik and R.~Mammone, ``Meta-neural networks that learn by learning,'' in
  \emph{Proc. Int. Joint Conf. on Neural Networks}, 1992, pp. 437--442.

\bibitem{schmidhuber1987evolutionary}
J.~Schmidhuber, ``Evolutionary principles in self-referential learning, or on
  learning how to learn: the meta-meta-... hook,'' Ph.D. dissertation, Institut
  f. Informatik, Tech. Univ. Munich, 1987.

\bibitem{zhao2019adaptive}
P.~Zhao, Y.~Zhang, M.~Wu, S.~C. Hoi, M.~Tan, and J.~Huang, ``Adaptive
  cost-sensitive online classification,'' \emph{{IEEE} Trans. Knowl. Data
  Eng.}, vol.~31, no.~2, pp. 214--228, 2019.

\bibitem{lake2015human}
B.~M. Lake, R.~Salakhutdinov, and J.~B. Tenenbaum, ``Human-level concept
  learning through probabilistic program induction,'' \emph{Science}, vol. 350,
  no. 6266, pp. 1332--1338, 2015.

\bibitem{koch2015siamese}
G.~Koch, R.~Zemel, and R.~Salakhutdinov, ``Siamese neural networks for one-shot
  image recognition,'' in \emph{Proc. Int. Conf. Machine Learning Workshop},
  2015.

\bibitem{Yu2019}
\BIBentryALTinterwordspacing
W.~Yu, X.~Lin, W.~Zhang, J.~Pei, and J.~A. McCann, ``Simrank*: effective and
  scalable pairwise similarity search based on graph topology,'' \emph{The VLDB
  Journal}, Jan 2019. [Online]. Available:
  \url{https://doi.org/10.1007/s00778-018-0536-3}
\BIBentrySTDinterwordspacing

\bibitem{mishra2018simple}
N.~Mishra, M.~Rohaninejad, X.~Chen, and P.~Abbeel, ``A simple neural attentive
  meta-learner,'' in \emph{Proc. Int. Conf. Learn. Representations}, 2018.

\bibitem{metz2018learning}
L.~Metz, N.~Maheswaranathan, B.~Cheung, and J.~Sohl-Dickstein, ``Learning
  unsupervised learning rules,'' in \emph{Proc. Int. Conf. Learn.
  Representations}, 2019.

\bibitem{gu2018meta}
J.~Gu, Y.~Wang, Y.~Chen, K.~Cho, and V.~O. Li, ``Meta-learning for low-resource
  neural machine translation,'' in \emph{Conference on Empirical Methods in
  Natural Language Processing (EMNLP)}, 2018.

\bibitem{schwartz2019repmet}
E.~Schwartz, L.~Karlinsky, J.~Shtok, S.~Harary, M.~Marder, S.~Pankanti,
  R.~Feris, A.~Kumar, R.~Giries, and A.~M. Bronstein, ``Repmet:
  Representative-based metric learning for classification and one-shot object
  detection,'' in \emph{Proc. {IEEE} Conf. Comput. Vis. Pattern Recognit.},
  2019.

\bibitem{allen2019infinite}
K.~R. Allen, E.~Shelhamer, H.~Shin, and J.~B. Tenenbaum, ``Infinite mixture
  prototypes for few-shot learning,'' in \emph{Proc. Int. Conf. Machine
  Learning}, 2019.

\bibitem{ppn}
L.~Liu, T.~Zhou, G.~Long, J.~Jiang, L.~Yao, and C.~Zhang, ``Prototype
  propagation networks ({PPN}) for weakly-supervised few-shot learning on
  category graph,'' in \emph{Proc. Int. Joint Conf. on AI}, 2019.

\bibitem{liu2019learning}
L.~Liu, T.~Zhou, G.~Long, J.~Jiang, and C.~Zhang, ``Learning to propagate for
  graph meta-learning,'' in \emph{Proc. Advances Neural Inf. Process. Syst.},
  2019, pp. 1037--1048.

\bibitem{tsoumakas2007multi}
G.~Tsoumakas and I.~Katakis, ``Multi-label classification: An overview,''
  \emph{International Journal of Data Warehousing and Mining (IJDWM)}, vol.~3,
  no.~3, pp. 1--13, 2007.

\bibitem{read2009classifier}
J.~Read, B.~Pfahringer, G.~Holmes, and E.~Frank, ``Classifier chains for
  multi-label classification,'' in \emph{Joint European Conference on Machine
  Learning and Knowledge Discovery in Databases}.\hskip 1em plus 0.5em minus
  0.4em\relax Springer, 2009, pp. 254--269.

\bibitem{read2014efficient}
J.~Read, L.~Martino, and D.~Luengo, ``Efficient monte carlo methods for
  multi-dimensional learning with classifier chains,'' \emph{Pattern
  Recognition}, vol.~47, no.~3, pp. 1535--1546, 2014.

\bibitem{read2015scalable}
J.~Read, L.~Martino, P.~M. Olmos, and D.~Luengo, ``Scalable multi-output label
  prediction: From classifier chains to classifier trellises,'' \emph{Pattern
  Recognition}, vol.~48, no.~6, pp. 2096--2109, 2015.

\bibitem{spolaor2013comparison}
N.~Spola{\^o}R, E.~A. Cherman, M.~C. Monard, and H.~D. Lee, ``A comparison of
  multi-label feature selection methods using the problem transformation
  approach,'' \emph{Electronic Notes in Theoretical Computer Science}, vol.
  292, pp. 135--151, 2013.

\bibitem{liu2018metric}
W.~Liu, D.~Xu, I.~Tsang, and W.~Zhang, ``Metric learning for multi-output
  tasks,'' \emph{{IEEE} Trans. Pattern Anal. Mach. Intell.}, vol.~41, no.~2,
  pp. 408--422, 2019.

\bibitem{liu2017making}
W.~Liu and I.~W. Tsang, ``Making decision trees feasible in ultrahigh feature
  and label dimensions,'' \emph{Journal of Machine Learning Research}, vol.~18,
  no.~1, pp. 2814--2849, 2017.

\bibitem{read2008multi}
J.~Read, B.~Pfahringer, and G.~Holmes, ``Multi-label classification using
  ensembles of pruned sets,'' in \emph{Proc. {IEEE} Int. Conf. on Data
  Mining}.\hskip 1em plus 0.5em minus 0.4em\relax IEEE, 2008, pp. 995--1000.

\bibitem{tsoumakas2009mining}
G.~Tsoumakas, I.~Katakis, and I.~Vlahavas, ``Mining multi-label data,'' in
  \emph{Data mining and knowledge discovery handbook}.\hskip 1em plus 0.5em
  minus 0.4em\relax Springer, 2009, pp. 667--685.

\bibitem{8413163}
S.~{Li}, Y.~{Jiang}, N.~V. {Chawla}, and Z.~{Zhou}, ``Multi-label learning from
  crowds,'' \emph{{IEEE} Trans. Knowl. Data Eng.}, vol.~31, no.~7, pp.
  1369--1382, July 2019.

\bibitem{heider2013multilabel}
D.~Heider, R.~Senge, W.~Cheng, and E.~H{\"u}llermeier, ``Multilabel
  classification for exploiting cross-resistance information in hiv-1 drug
  resistance prediction,'' \emph{Bioinformatics}, vol.~29, no.~16, pp.
  1946--1952, 2013.

\bibitem{silla2011survey}
C.~N. Silla and A.~A. Freitas, ``A survey of hierarchical classification across
  different application domains,'' \emph{Data Mining and Knowledge Discovery},
  vol.~22, no. 1-2, pp. 31--72, 2011.

\bibitem{gordon1987review}
A.~D. Gordon, ``A review of hierarchical classification,'' \emph{Journal of the
  Royal Statistical Society: Series A (General)}, vol. 150, no.~2, pp.
  119--137, 1987.

\bibitem{wehrmann2018hierarchical}
J.~Wehrmann, R.~Cerri, and R.~Barros, ``Hierarchical multi-label classification
  networks,'' in \emph{Proc. Int. Conf. Machine Learning}, 2018, pp.
  5225--5234.

\bibitem{dumais2000hierarchical}
S.~Dumais and H.~Chen, ``Hierarchical classification of web content,'' in
  \emph{Proceedings of the International ACM SIGIR conference on Research and
  development in Information Retrieval}.\hskip 1em plus 0.5em minus 0.4em\relax
  ACM, 2000, pp. 256--263.

\bibitem{sun2001hierarchical}
A.~Sun and E.-P. Lim, ``Hierarchical text classification and evaluation,'' in
  \emph{Proc. {IEEE} Int. Conf. on Data Mining}.\hskip 1em plus 0.5em minus
  0.4em\relax IEEE, 2001, pp. 521--528.

\bibitem{gauch1981hierarchical}
H.~G. Gauch~Jr and R.~H. Whittaker, ``Hierarchical classification of community
  data,'' \emph{The Journal of Ecology}, pp. 537--557, 1981.

\bibitem{8333767}
Q.~{Zhang}, C.~{Shi}, Z.~{Niu}, and L.~{Cao}, ``Hcbc: A hierarchical case-based
  classifier integrated with conceptual clustering,'' \emph{{IEEE} Trans.
  Knowl. Data Eng.}, vol.~31, no.~1, pp. 152--165, Jan 2019.

\bibitem{yang2019hierarchical}
Y.~Yang, Y.~Duan, X.~Wang, Z.~Huang, N.~Xie, and H.~T. Shen, ``Hierarchical
  multi-clue modelling for poi popularity prediction with heterogeneous tourist
  information,'' \emph{{IEEE} Trans. Knowl. Data Eng.}, vol.~31, no.~4, pp.
  757--768, 2019.

\bibitem{lyu2019supergraph}
B.~Lyu, L.~Qin, X.~Lin, L.~Chang, and J.~X. Yu, ``Supergraph search in graph
  databases via hierarchical feature-tree,'' \emph{{IEEE} Trans. Knowl. Data
  Eng.}, vol.~31, no.~2, pp. 385--400, 2019.

\bibitem{Ouyang18}
\BIBentryALTinterwordspacing
D.~Ouyang, L.~Qin, L.~Chang, X.~Lin, Y.~Zhang, and Q.~Zhu, ``When hierarchy
  meets 2-hop-labeling: Efficient shortest distance queries on road networks,''
  in \emph{Proceedings of the 2018 International Conference on Management of
  Data}, ser. SIGMOD '18.\hskip 1em plus 0.5em minus 0.4em\relax New York, NY,
  USA: ACM, 2018, pp. 709--724. [Online]. Available:
  \url{http://doi.acm.org/10.1145/3183713.3196913}
\BIBentrySTDinterwordspacing

\bibitem{ren2018meta}
M.~Ren, E.~Triantafillou, S.~Ravi, J.~Snell, K.~Swersky, J.~B. Tenenbaum,
  H.~Larochelle, and R.~S. Zemel, ``Meta-learning for semi-supervised few-shot
  classification,'' in \emph{Proc. Int. Conf. Learn. Representations}, 2018.

\bibitem{li2019large}
A.~Li, T.~Luo, Z.~Lu, T.~Xiang, and L.~Wang, ``Large-scale few-shot learning:
  Knowledge transfer with class hierarchy,'' in \emph{Proceedings of the IEEE
  Conference on Computer Vision and Pattern Recognition}, 2019, pp. 7212--7220.

\bibitem{yao2019hierarchically}
H.~Yao, Y.~Wei, J.~Huang, and Z.~Li, ``Hierarchically structured
  meta-learning,'' 2019.

\bibitem{valentini2011true}
G.~Valentini, ``True path rule hierarchical ensembles for genome-wide gene
  function prediction,'' \emph{IEEE/ACM Transactions on Computational Biology
  and Bioinformatics (TCBB)}, vol.~8, no.~3, pp. 832--847, 2011.

\bibitem{cesa2012synergy}
N.~Cesa-Bianchi, M.~Re, and G.~Valentini, ``Synergy of multi-label hierarchical
  ensembles, data fusion, and cost-sensitive methods for gene functional
  inference,'' \emph{Machine Learning}, vol.~88, no. 1-2, pp. 209--241, 2012.

\bibitem{yu2015predicting}
G.~Yu, H.~Zhu, C.~Domeniconi, and J.~Liu, ``Predicting protein function via
  downward random walks on a gene ontology,'' \emph{BMC bioinformatics},
  vol.~16, no.~1, p. 271, 2015.

\bibitem{yu2018newgoa}
G.~Yu, G.~Fu, J.~Wang, and Y.~Zhao, ``Newgoa: Predicting new go annotations of
  proteins by bi-random walks on a hybrid graph,'' \emph{IEEE/ACM Transactions
  on Computational Biology and Bioinformatics (TCBB)}, vol.~15, no.~4, pp.
  1390--1402, 2018.

\bibitem{gholami2017probabilistic}
B.~Gholami and V.~Pavlovic, ``Probabilistic temporal subspace clustering,'' in
  \emph{Proc. {IEEE} Conf. Comput. Vis. Pattern Recognit.}, 2017, pp.
  3066--3075.

\bibitem{vaswani2017attention}
A.~Vaswani, N.~Shazeer, N.~Parmar, J.~Uszkoreit, L.~Jones, A.~N. Gomez,
  {\L}.~Kaiser, and I.~Polosukhin, ``Attention is all you need,'' in
  \emph{Proc. Advances Neural Inf. Process. Syst.}, 2017, pp. 5998--6008.

\bibitem{rusu2018meta}
A.~A. Rusu, D.~Rao, J.~Sygnowski, O.~Vinyals, R.~Pascanu, S.~Osindero, and
  R.~Hadsell, ``Meta-learning with latent embedding optimization,'' in
  \emph{Proc. Int. Conf. Learn. Representations}, 2018.

\bibitem{nichol2018first}
A.~Nichol, J.~Achiam, and J.~Schulman, ``On first-order meta-learning
  algorithms,'' \emph{arXiv preprint arXiv:1803.02999}, 2018.

\bibitem{GroupNorm2018}
Y.~Wu and K.~He, ``Group normalization,'' in \emph{Proc. Eur. Conf. Comput.
  Vis.}, 2018, pp. 3--19.

\bibitem{ioffe2015batch}
S.~Ioffe and C.~Szegedy, ``Batch normalization: Accelerating deep network
  training by reducing internal covariate shift,'' in \emph{Proc. Int. Conf.
  Machine Learning}, 2015, pp. 448--456.

\bibitem{kingma2015adam}
D.~P. Kingma and J.~Ba, ``Adam: A method for stochastic optimization,'' in
  \emph{Proc. Int. Conf. Learn. Representations}, 2015.

\bibitem{meta-mem-aug}
A.~Santoro, S.~Bartunov, M.~Botvinick, D.~Wierstra, and T.~Lillicrap,
  ``Meta-learning with memory-augmented neural networks,'' in \emph{Proc. Int.
  Conf. Machine Learning}, 2016, pp. 1842--1850.

\bibitem{maaten2008visualizing}
L.~v.~d. Maaten and G.~Hinton, ``Visualizing data using t-sne,'' \emph{Journal
  of Machine Learning Research}, vol.~9, no.~11, pp. 2579--2605, 2008.

\end{thebibliography}

% biography section
% 
% If you have an EPS/PDF photo (graphicx package needed) extra braces are
% needed around the contents of the optional argument to biography to prevent
% the LaTeX parser from getting confused when it sees the complicated
% \includegraphics command within an optional argument. (You could create
% your own custom macro containing the \includegraphics command to make things
% simpler here.)
%\begin{IEEEbiography}[{\includegraphics[width=1in,height=1.25in,clip,keepaspectratio]{mshell}}]{Michael Shell}
% or if you just want to reserve a space for a photo:
\vspace{-3em}
\begin{IEEEbiography}[{\includegraphics[width=1in,height=1.25in,clip,keepaspectratio]{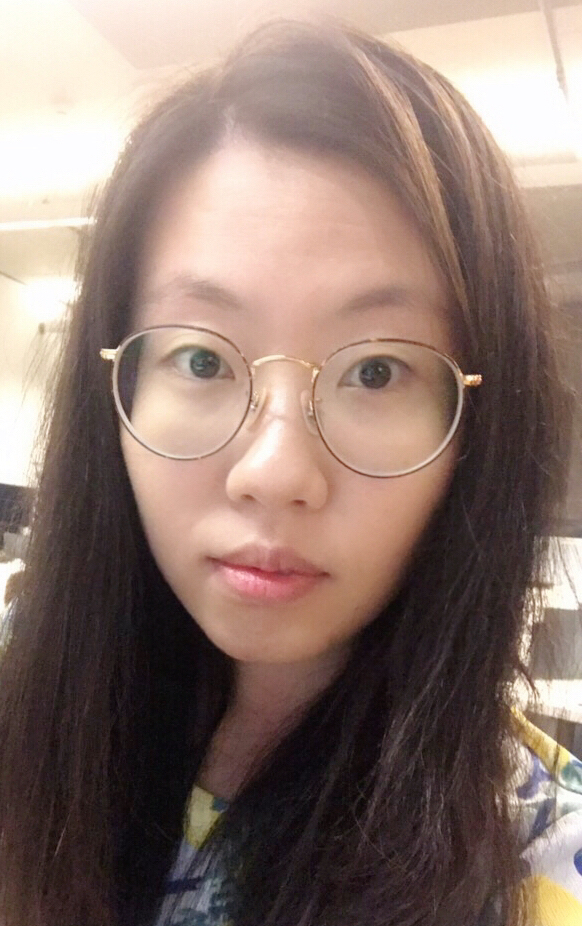}}]{Lu Liu}
received her bachelor’s degree from South China University of Technology (SCUT), Guangzhou, China, in 2017. 
% She worked with Prof. Fethi Rabhi at University of New South Wales, Australia, as a research exchange student during 2016 to 2017. 
She is currently pursuing the Ph.D. at the Center for Artificial Intelligence, University of Technology Sydney (UTS), Australia.
Her current research interests include deep learning, machine learning, few-shot learning and meta-learning.
\end{IEEEbiography}
\vspace{-3em}
\begin{IEEEbiography}[{\includegraphics[width=1in,height=1.25in,clip,keepaspectratio]{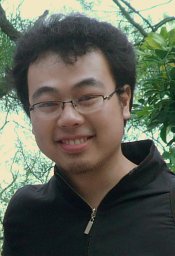}}]{Tianyi Zhou}
is a Ph.D student of Paul G. Allen School of Computer Science and Engineering at University of Washington, Seattle. His research covers several topics of machine learning, natural language processing, statistics and data analysis. He has published 30+ papers at top conferences and journals including NeurIPS, ICML, ICLR, AISTATS, ACM SIGKDD, IEEE ICDM, AAAI, IJCAI, IEEE ISIT, Machine Learning Journal (Springer), DMKD (Springer), IEEE TIP, IEEE TNNLS, etc. He is the recipient of the best student paper award at IEEE ICDM 2013.
\end{IEEEbiography}
\vspace{-3em}
\begin{IEEEbiography}[{\includegraphics[width=1in,height=1.25in,clip,keepaspectratio]{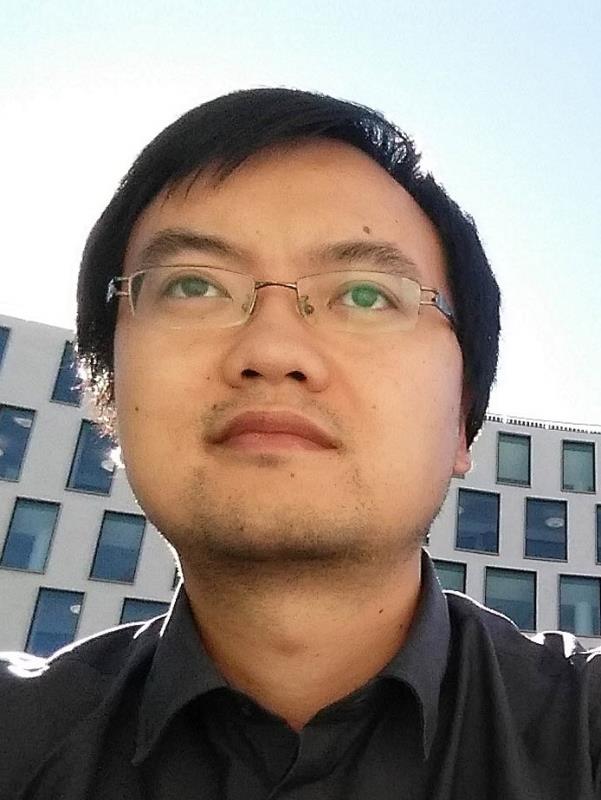}}]{Guodong Long}
received his PhD degree from the University of Technology Sydney (UTS), Australia, in 2014. He is a senior lecturer at the Centre for Artificial Intelligence (CAI), Faculty of Engineering and IT, UTS. His research focuses on data mining, machine learning, and natural language processing. He has more than 40 research papers published on top-tier journals and conferences, including IEEE TPAMI, TCYB, TKDE, ICLR, AAAI, IJCAI, and ICDM.
\end{IEEEbiography}
\vspace{-3em}
\begin{IEEEbiography}[{\includegraphics[width=1in,height=1.25in,clip,keepaspectratio]{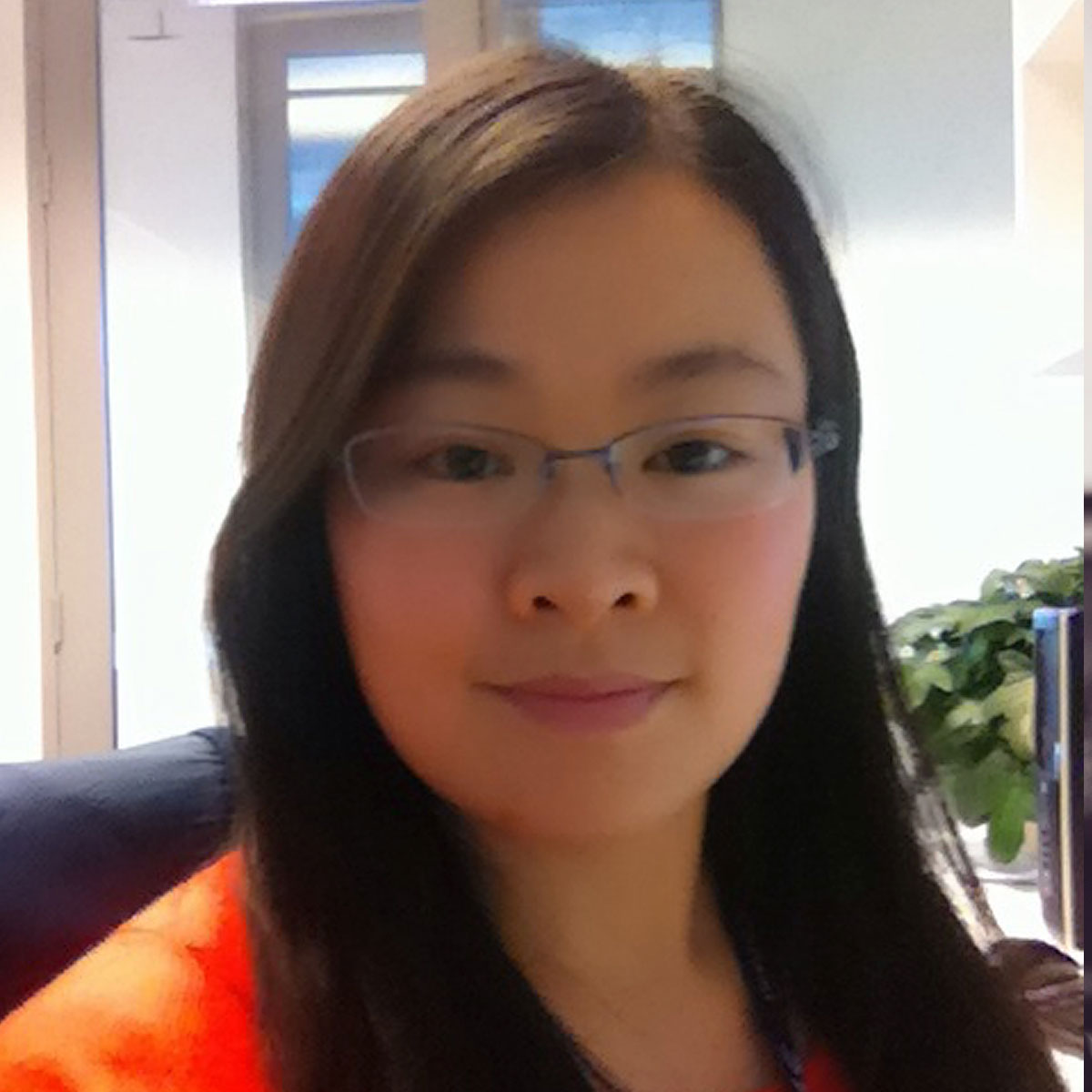}}]{Jing Jiang} received her PhD degree from the University of Technology Sydney (UTS), Australia in 2015. She is currently a Lecturer at the Centre for Artificial Intelligence, Faculty of Engineering and IT, UTS. Her research interest lies in data mining and machine learning applications with the focuses on deep reinforcement learning and sequential decision-making. She has more than 20 research papers published on top-tier journals and conferences including AAAI, IJCAI and CCGrid.
\end{IEEEbiography}
\vspace{-3em}
\begin{IEEEbiography}[{\includegraphics[width=1in,height=1.25in,clip,keepaspectratio]{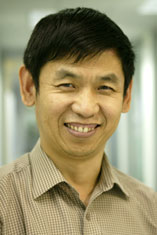}}]{Chengqi Zhang}
has been appointed as a Distinguished Professor at the University of Technology Sydney from 27 February 2017 to 26 February 2022, an Executive Director UTS Data Science from 3 January 2017 to 2 January 2021, an
Honorary Professor at the University of Queensland from 1 January 2015 to 31 December 2017,
an Adjunct Professor at the University of New
South Wales from 20 March 2017 to 20 March
2020, and a Research Professor of Information
Technology at UTS from 14 December 2001.
In addition, he has been selected as the Chairman of the Australian
Computer Society National Committee for Artificial Intelligence since
November 2005, and the Chairman of IEEE Computer Society Technical
Committee of Intelligent Informatics (TCII) since June 2014.
% Prof. Zhang obtained his Bachelor degree from Fudan University in
% March 1982, Master degree from Jilin University in March 1985, PhD
% degree from The University of Queensland in October 1991, followed
% by a Doctor of Science from Deakin University in October 2002, all in
% Computer Science.
\end{IEEEbiography}

% You can push biographies down or up by placing
% a \vfill before or after them. The appropriate
% use of \vfill depends on what kind of text is
% on the last page and whether or not the columns
% are being equalized.

% \vfill

% Can be used to pull up biographies so that the bottom of the last one
% is flush with the other column.
%\enlargethispage{-5in}

% that's all folks
\end{document}